\documentclass{article} 
\usepackage{iclr2027_conference,times}

\usepackage{graphicx}
\usepackage{booktabs}
\usepackage{hyperref}
\usepackage{url}
\usepackage{multirow}
\usepackage{array}
\usepackage{amsmath}
\usepackage{amsfonts}
\usepackage{amssymb}
\usepackage{amsthm}
\usepackage{multicol}
\usepackage{caption}
\usepackage{subcaption}
\usepackage{tabularx}
\usepackage{placeins}

\usepackage{CJKutf8}
\usepackage{array}
\usepackage[main=english,russian]{babel}
\usepackage{makecell}
\usepackage{fontawesome5}
\usepackage{xcolor}

\usepackage[capitalize]{cleveref}
\crefformat{section}{\S#2#1#3}
\crefformat{subsection}{\S#2#1#3}
\crefformat{subsubsection}{\S#2#1#3}

\newif\ifcommentson
\commentsonfalse 
\newcommand{\draftonly}[1]{#1}
\renewcommand{\draftonly}[1]{}  
\newcommand{\draftcomment}[1]{\draftonly{#1}}
\newcommand{\todo}[1]{\draftcomment{\textcolor{red}{\small [TODO: #1]}}}
\newcommand{\sarah}[1]{\draftcomment{\textcolor{orange}{\small [Sarah: #1]}}}

\newcommand{\stephen}[1]{\draftcomment{\textcolor{purple}{\small [stephen: #1]}}}

\definecolor{revisecolor}{rgb}{0.6,0,0}
\newif\ifrevisions
\revisionstrue      
\revisionsfalse   
\newcommand{\revise}[1]{\ifrevisions{\color{revisecolor}#1}\else#1\fi}

\title{On the Interplay between Backtracking and Latent Confidence Directions in Reasoning Models}

\author{Stephen Cheng$^1$,\; 
Zahra Sodagar$^1$,\; 
Sandeep Soni$^2$,\; 
Dinesh Manocha$^1$,\; 
Sarah Wiegreffe$^1$ \\
$^1$University of Maryland, $^2$Emory University \\
\texttt{scheng03@umd.edu}
}

\preprint
\begin{document}

\maketitle

\begin{abstract}
Reasoning models deliberate over multiple answer proposals, often resulting in excessively long traces due to backtracking, even when the correct answer has already been reached
(\revise{Across three 32B models}, $\sim$87\% of rollouts contain 1 unique answer yet still make $\sim$8 answer proposals on average).
Given this potentially ``decorative" chain of thought (CoT) reasoning, can we look to the model internals to determine if the model will backtrack or commit to an answer proposal?
Operating under the hypothesis that models maintain an internal representation of their uncertainty,
we seek to extract this representation and characterize its relationship with backtracking behavior at answer proposals.
By querying the model's verbalized confidence following answer proposals throughout a response trace,
we extract a linear direction for latent confidence
that separates high and low verbalized confidence.
This direction can steer verbalized confidence as expected, but it surprisingly can also steer the next token probability of backtracking, performing on par with a linear direction extracted directly from backtracking examples. 
We then test the direction under progressively more realistic interventions. Steering continuously from \revise{a mid-trace} answer proposal modulates response length ($\sim$-1.3k -- $\sim$+2.5k tokens) and the number of answer proposals (\revise{-4.18} -- \revise{+4.75}) in both directions, with minimal cost to accuracy.
Applying the confidence direction from the start of generation, where proposal positions are unknown, still shortens traces at comparable accuracy, and its effect surfaces in the linguistic confidence of proposals.
Overall, our study reveals the intricate interplay between a model's internal confidence and its proclivity to backtrack or commit to an answer.

\end{abstract}

\section{Introduction}
Reasoning large language models (LLMs) \citep{Guo_2025, yang2025qwen3technicalreport} solve difficult math \citep{shao2024deepseekmathpushinglimitsmathematical}, science, and programming problems \citep{Jiang_2026} by generating long chains of thought (CoT) \citep{NEURIPS2022_9d560961} to reason through the problem before arriving at a final answer.
A defining characteristic of these traces is backtracking \citep{gandhi2025cognitive}, where the model chooses to revise its work during its CoT. Although backtracking behavior can improve performance by helping the model to fix its mistakes and explore alternate possible solutions \citep[][\emph{i.a.}]{muennighoff-etal-2025-s1, kim2025astroteachinglanguagemodels, yang2025demystifying, shen2025satori, ai2025rethinkingreflectionpretraining, tian2025thinktwiceenhancingllm, cai2026how}, it also often leads to excessively long or redundant outputs \citep{chen2025do, su2025underthinkingoverthinkingempiricalstudy, wang_wait_2025}.

We study backtracking behavior at \emph{answer proposals}, sentences in which the model arrives at a candidate answer and could plausibly stop reasoning. 
A model may propose an answer to a question multiple times before committing to a final answer, though we observe that models rarely change their answer value from the initial proposal.
Across five benchmarks and three open reasoning models, 84--89\% of rollouts propose only a single unique answer, yet the model proposes that answer 5-9 times on average before terminating reasoning (\autoref{tab:unique-vs-proposed-summary}). 

This raises the question: \emph{when a reasoning model proposes an answer, what factors in the internal representations determine whether it commits or backtracks?}
From qualitative inspection, a model's linguistic confidence varies across a reasoning trace, and typically increases steadily as the model nears answer commitment. 
Thus, a natural hypothesis is that reasoning models maintain an internal representation of confidence in their reasoning progress, which not only affects linguistic confidence but also drives answer commitment.
We aim to extract an approximation of this internal confidence and measure its causal role in controlling answer commitment.
\stephen{maybe a linking sentence here}
\sarah{also very long PP; split in two?}

Uncertainty quantification \citep{shorinwa2025surveyuncertaintyquantificationlarge} work demonstrates that LLMs' verbalized confidence \citep{yoon2025reasoning} is a calibrated confidence measure for answer correctness.
We query models' verbalized confidence at each answer proposals in the CoT rollout, a sort of ``textual probe'', and fit a contrastive linear direction on the model's residual stream representations 
that separates high from low verbalized confidence.
Applying this direction at different points in the CoT not only steers verbalized confidence, but also the model's likelihood of backtracking, indicating that these two tasks share some internal structure in activation space \sarah{too strong?} (\cref{sec:4}).
Interestingly, the cross-task transfer does not hold in the reverse direction: a linear direction that separates backtracking vs. answer commitment effectively steers backtracking probability but cannot steer verbalized confidence. 
Finally, to better understand this confidence direction's causal effect on CoTs, we verify that it both reduces (for positive steering) and increases (for negative steering) the number of answer proposals and overall CoT length as expected (\cref{sec:5postmarker}, \cref{sec:5steer_scratch})-- for example, positive steering following mid-trace answer proposals reduces additional answer proposals by \revise{4.18} proposals and response length by \revise{1323} tokens, averaged across three models.
\revise{Qualitatively, the effect of confidence steering is also reflected in the degree of confident language used by the model (\cref{sec:5steer_qualitative})}. 
Our results provide evidence that latent confidence signals can be 1) extracted by prompting a model to elicit their verbalized confidence and 2) accumulated to approximate a linear confidence direction that can be used to 3) effectively steer answer commitment.\todo{add something more concrete \#s here per Dinesh's comment}

To summarize, our contributions are:\sarah{@Stephen please check this, esp \#3 ``including when'' and \#4 (I think we can only claim it's comparable?)}
\begin{itemize}
  \item \textbf{Answer proposal redundancy.} Across three 32B reasoning LLMs, only 11-16\% of rollouts contain more than one unique answer despite many answer proposals. This motivates our study on answer proposal positions and their relationship with model confidence (\cref{sec:3}).
  \item \textbf{Cross-task transfer of latent confidence}. A confidence direction extracted from verbalized confidence reports causally steers backtracking probability after answer proposals, matching a direction fit on commitment itself, while the reverse transfer fails (\cref{sec:4}). Latent confidence more reliably steers commitment behavior than CoT-level confidence (\cref{sec:5cot_vs_steer}).
  \item \textbf{Trace-level control with confidence steering.} When applied after answer proposals (\cref{sec:5postmarker}) or at inference time (\cref{sec:5steer_scratch}), continuous confidence steering bidirectionally controls the number of answer proposals, response length, and the linguistic confidence of the CoT with minor or no cost to accuracy.
\end{itemize}

\section{Related Work}
\paragraph{Interpreting Reasoning and Backtracking}
A growing body of work studies how reasoning models internally represent and control behaviors such as backtracking.
One line of work uses sparse dictionary learning to find unsupervised linear features associated with reasoning patterns and interpret them post-hoc
\citep{venhoff2026basemodelsknowreason, zhang2026fantastic,troitskii-etal-2025-internal}.
Separately, \citet{zhang2025reasoning} and \citet{boppana2026reasoningtheaterdisentanglingmodel} show that probes can decode answers or predict answer accuracy before models finish their reasoning trace, and use these probes as a confidence proxy for early exiting. 
\revise{
A similar line of work extracts linear directions and steers along them to reduce reasoning length \citep{eisenstadt2025overclockingllmreasoningmonitoring, azizi-etal-2026-activation}.
} 
\citet{venhoff2025understandingreasoningthinkinglanguage} extract steering vectors for reasoning behaviors and evaluate them on how well they control the frequency of the behavior they were trained on.
Our work differs in two ways. First, rather than fitting a direction on the behavior we aim to steer, we derive it from the model's own verbalized confidence and demonstrate its generalization to a different behavior, answer commitment. Second, we evaluate the confidence direction on various task-level metrics (probability of backtracking, number of answer proposals, response length, accuracy) across RL-trained and distilled reasoning models.

\paragraph{Verbalized Uncertainty Quantification and Internal Confidence}
Uncertainty quantification (UQ) \citep{geng-etal-2024-survey} is concerned with obtaining a confidence metric that is calibrated with the model's probability of answer correctness.
Prompting the model to verbalize its uncertainty \citep{lin2022teachingmodelsexpressuncertainty, tian-etal-2023-just,yang2026verbalizedconfidencescoresllms} is a widely studied approach.
While past research questions the reliability of verbalized uncertainty \citep{groot2024overconfidencekeyverbalizeduncertainty, ICLR2024_6733cf15},
compared to token-level \citep{duan2024shiftingattentionrelevancepredictive} or resampling \citep{lin2024generatingconfidenceuncertaintyquantification} approaches, 
recent work finds that modern reasoning LLMs' verbalized confidence is well calibrated  \citep{yoon2025reasoning}.
Rather than predicting \textit{final} answer correctness, we leverage verbalized confidence prompting to extract a latent confidence direction and measure its effects across \textit{intermediate} answer proposals.
\citet{kumaran2026llmscomputeverbalconfidence} investigate how models compute verbalized confidence by analyzing hidden state interactions, finding that verbalized confidence reflects internal self-evaluation beyond token log probabilities.
\revise{
Similar work assesses if and how LLMs leverage internal confidence on curated synthetic experiments \citep{ackerman2026evidence} or non-CoT MCQA \citep{Kumaran2026}.}
Building on this, we measure how internal confidence influences long CoT reasoning behavior.

\section{Methodology}
\label{sec:3}

\begin{table}[t]
  \centering
  \scriptsize
  \caption{
  We pool the rollouts from AIME 2026, GSM8K, OMEGA-500, GPQA-D, MMLU-Pro and calculate the \% of rollouts that have $n$ unique answers. For each $n$, we record the mean $\pm$ std number of proposal markers. Note that std is high because we average over five datasets. The All column computes over all rollouts. See \autoref{tab:unique-vs-proposed} for per-dataset results.}
  \label{tab:unique-vs-proposed-summary}
  \begin{tabular}{lr rr@{\,$\pm$\,}l rr@{\,$\pm$\,}l rr@{\,$\pm$\,}l r@{\,$\pm$\,}l}
    \toprule
    \multirow{2}{*}{Model} & \multirow{2}{*}{Total \#}
      & \multicolumn{3}{c}{1 unique} & \multicolumn{3}{c}{2 unique}
      & \multicolumn{3}{c}{3+ unique} & \multicolumn{2}{c}{All} \\
    \cmidrule(lr){3-5} \cmidrule(lr){6-8} \cmidrule(lr){9-11} \cmidrule(lr){12-13}
     & rollouts & \% rollouts & \multicolumn{2}{c}{proposals}
       & \% rollouts & \multicolumn{2}{c}{proposals}
       & \% rollouts & \multicolumn{2}{c}{proposals}
       & \multicolumn{2}{c}{proposals} \\
    \midrule
    Olmo 3 32B  & 3202 & 84.4 & 8.6 & 6.5 & 11.7 & 16.8 & 12.2 & 3.8 & 26.6 & 21.6 & 10.2 & 9.4 \\
    Qwen 3 32B   & 3197 & 89.3 & 6.0 & 4.9 &  8.0 & 11.3 &  7.7 & 2.7 & 23.5 & 16.4 &  6.9 & 6.6 \\
    Distill 32B & 3025 & 87.6 & 4.7 & 4.3 &  9.3 & 11.2 & 10.6 & 3.1 & 16.7 & 13.4 &  5.7 & 6.3 \\
    \bottomrule
  \end{tabular}
\end{table}

\subsection{Obtaining Answer Proposals}
\label{sec:3answerproposals}

\paragraph{Answer Proposal Definition} A reasoning trace alternates between computation blocks, where the model works towards an answer, and answer proposal sentences. After each intermediate answer proposal, the model typically begins the next computation block expressing its intent to revise or verify its work.
While backtracking may occur within computation blocks, we study backtracking vs. commitment behavior immediately following answer proposals, since answer proposals serve as a natural point to break the trace into blocks.

We define a sentence to be an \textit{answer proposal} if it satisfies two conditions:
1) it contains an answer value, and 2) if read up to that point, it is plausible that the model can commit to the answer and end reasoning within the next couple sentences.
A sentence is an answer proposal only if the model is arriving at an answer (e.g., ``Therefore, I think the answer is \textit{X}."), not simply questioning if the answer is correct (e.g., ``Let me make sure \textit{X} is the correct answer.") or hedging a guess (e.g., ``Maybe the answer is \textit{X}.").

\paragraph{Extracting Answer Proposals} We identify answer proposal positions within the reasoning traces (i.e., within the \texttt{<think>} \texttt{</think>} tags) across sampled reasoning model generations using an LLM judge pipeline. Compared against golden label human annotations by the authors, the pipeline achieves an average precision of 0.89.
After the judge pipeline, we manually review rollouts to ensure that the last answer proposal is always selected.
See App. \ref{app:judge} for additional details. For the rest of the paper, we refer to the last token of an answer proposal sentence as a \emph{marker}. Given $n$ markers in a rollout, we denote the first $n-1$ as mid-trace markers and the $n$th marker as the final\sarah{$\leftarrow$ true that this is always the same as the final answer? if so, mention that}.

We perform this process on sampled rollouts generated by Qwen 3 32B \citep{yang2025qwen3technicalreport}, Olmo 3 32B Think \citep{olmo2025olmo3}, and DeepSeek R1 Distill Qwen 32B \citep{deepseekai2025deepseekr1incentivizingreasoningcapability}
\footnote{Subsequently abbreviated as ``Qwen 3 32B'', ``Olmo 3 32B'', and ``Distill 32B'', respectively.}
on the AIME 2026 \citep{aime26}, GSM8K \citep{cobbe2021trainingverifierssolvemath}, OMEGA-500 \citep{sun2024omega}, GPQA Diamond \citep{rein2023gpqagraduatelevelgoogleproofqa}, and MMLU-Pro \citep{wang2024mmluprorobustchallengingmultitask} datasets. The first three are mathematical reasoning datasets and the latter two are science and general knowledge multiple-choice questions. 
We sample up to 200 prompts from each dataset and sample 4 reasoning traces per prompt from each model with max response length of 32,768 tokens using a temperature=0.6 and top-p=0.95.\footnote{We only retain traces where the model reached a final prediction within the max. reasoning budget.} 
\autoref{tab:unique-vs-proposed-summary} records the number of unique answers and number of proposal markers per model and dataset.
Averaged across models and datasets, models provide one unique answer around 87\% of the time and 8 proposals on average, indicating a high propensity to double check and verify their proposed answers. \revise{This complements \citet{chen2025do}, who count the number of answer proposals; we also count unique answers.}

\subsection{Extracting Latent Confidence}
\label{sec:3dimconstruction}

We seek to measure what causes a model to reconsider or commit at an answer proposal. We hypothesize that reasoning models maintain a representation of their confidence surrounding their progress on the problem, and that they typically commit once they are confident enough.
As the model works through and double checks its answers, it gradually increases its confidence, thus leading to answer commitment.
In this section, we detail how we approximate latent confidence as a linear direction.

\paragraph{Querying Confidence}
At each selected answer proposal marker, we append the prompt,
\texttt{\small My confidence in this answer, on a scale from 0 to 9, is }, and extract the expected confidence from the probabilities of the digit tokens.
In this way, we let the model evaluate its confidence in its answer, and verbalize it as an ordinal value.
Across all models and datasets, verbalized confidence and binary answer commitment have a modest pooled Spearman $\rho=0.15$, and a within-rollout $\rho=0.22$, as shown in \autoref{tab:vc_commit_correlations}.
\revise{\footnote{\revise{Due to the imbalanced ratio of binary labels, average pooled $\rho_{\max} = 0.59$ and within-rollout $\rho_{\max}=0.57$.}}}
\footnote{UQ work often also uses token-level entropy as an uncertainty measurement for the model's final answer \citep{fadeeva-etal-2024-fact, grunefeld2026tracinguncertaintylanguagemodel}. We measure the sentence-level entropy of mid-trace and final answer proposals and find no significant relationship.}
We view verbalized confidence as a surface-level approximation of the model's underlying confidence, and ask if we can extract a latent direction from the verbalized confidence data that steers other confidence-related traits.

\paragraph{Extracting confidence direction}
We divide the answer proposal markers into train, val, and test splits (0.4/0.2/0.4) such that all markers from each trace belong to the same split and the traces within each split differ in accuracy by no more than 5\%.
To extract the confidence direction, we assign markers as high or low confidence based on the train split's median verbalized confidence value, and obtain the Difference-in-Means (DiM) vector \citep{rimsky-etal-2024-steering, belrose2023diffmeansworstcaseoptimal}.
We extract
\begin{equation*}
    v_\text{conf}^s = \frac{1}{|M^+|} \sum_{p \in M^+} \mathbf{h}^s(p)
    - \frac{1}{|M^-|} \sum_{q \in M^-} \mathbf{h}^s(q)
\end{equation*}
where $M^+, M^-$ are the high and low confidence markers, and $h^s$ is the site $s$ representation at the last token position of the verbalized confidence prompt. To account for the difference in the number of available traces across different datasets, we sample from the train split so that $|M^+|=|M^-| = 100$. At inference-time, we add or subtract $v_\text{conf}^s$ at the post-attention or post-MLP residual states. Formally, given a model with $N$ transformer layers, the model has $2N$ steering sites at this layer of granularity; we thus obtain $2N$ DiM directions.

\begin{figure}[t]
    \centering
    \includegraphics[width=0.95\linewidth]{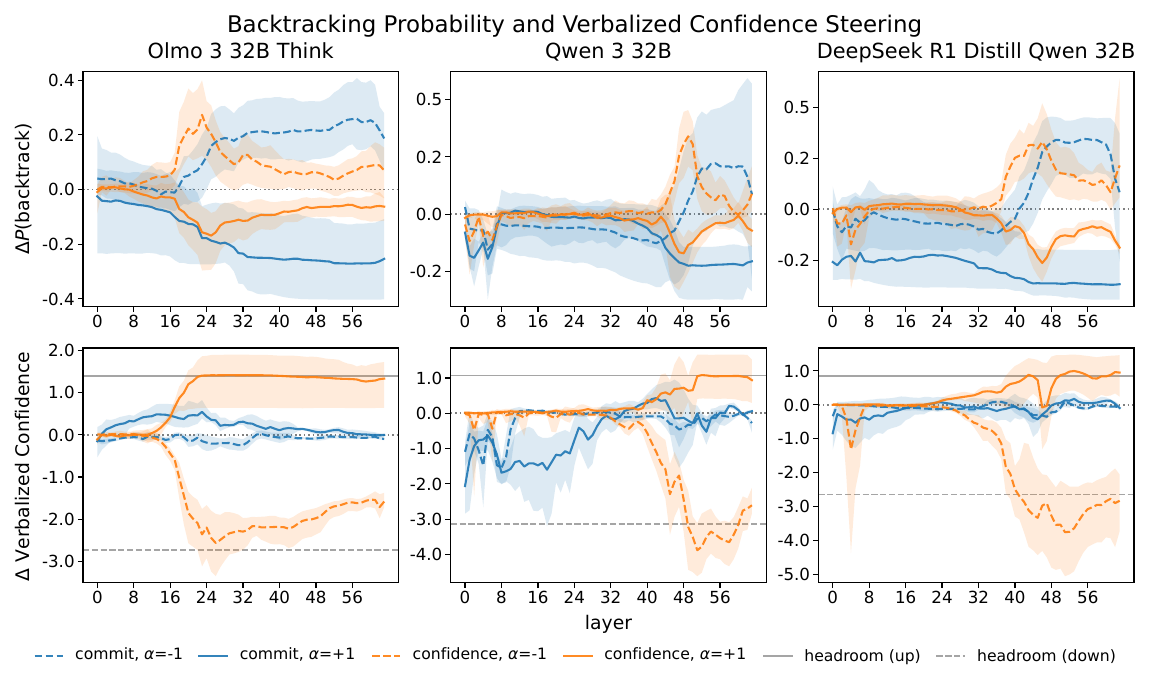}
    \caption{\textbf{Top}: Steering a model's probability of emitting a backtracking token as the next token after an answer proposal. \textbf{Bottom}: Steering verbalized confidence. Results are averaged across AIME 2026, GSM8K, Omega-500, GPQA-D, and MMLU-Pro. See \autoref{fig:st_commit_steer_all} and \autoref{fig:st_conf_steer_all} for each dataset's validation sweeps.}
    \label{fig:st_steer}
\end{figure}

\sarah{comment for a later date when we have more time, but the exact token positions we're steering at and with which vectors is a bit hard to track and would prob benefit from a figure. we also don't have a "figure 1", which we can consider adding later.}
\paragraph{Steering verbalized confidence (VC)} For each model, we computed five sets of $v_\text{conf}$, one from each dataset's train split. 
To verify that our confidence directions can steer on the task in which they were fit, we test $v_\text{conf}$ on steering VC at the last token position of the VC prompt via activation addition \citep{turner2023steering}, where positive steering should increase VC and negative steering should decrease VC.
We perform a validation sweep across the steering sites with $v_\text{conf}$ for each dataset.
Details on the setup are in App. \ref{app:st_steer_details}.
The bottom row in \autoref{fig:st_steer} shows the averaged $\Delta$ VC across all datasets for $\alpha=\pm1$ in orange.
For context, we also plot the difference between the marker's unsteered VC and the 2nd and 98th percentile VC values, averaged across all markers in the dataset. This headroom represents the approximate upper bound for positive and negative steering, based on the observed VC values that the model naturally emits.
As expected, we find that the confidence direction easily steers VC, and it saturates in performance after reaching early-middle layers in Olmo 3 32B, and middle-late layers in Qwen 3 32B and Distill 32B.

\paragraph{Confidence direction robustness}
The cosine similarity across all datasets' $v_\text{conf}$ per model is at least 0.85, implying that a general confidence direction is shared across datasets.
To ensure that the confidence direction is not encoding a specific numerical difference, we query verbalized confidence using two alternative prompts: \texttt{\small My confidence in this answer, on a scale from 1 to 100, is } and \texttt{\small My confidence in this answer, chosen from "low", "medium", "high", is }.
We validate that the confidence direction is robust to the phrasing of the verbalized prompt in App. \ref{app:confidence_consistency}.
Going forward, we use the confidence direction extracted from the range 9 prompt.

\section{Steering Backtracking Probability with Confidence}

\label{sec:4}

\subsection{Steering Backtracking Token Probability}
\label{sec:4st_steer}
As shown in \autoref{tab:backtrack-freq}, a mid-trace answer proposal is followed by a backtracking word (``Wait", ``But", ``Hmm", etc.) at least 50\% of the time across all models, pooled across datasets.
On the other hand, a final answer proposal is followed by a backtracking word $\lesssim$2\% of the time.
Thus, we treat the next token probability of predicting a backtracking token, $p_\text{bt}$, as a proxy for answer proposal commitment, and test the confidence representations on steering $p_\text{bt}$ at the proposal marker.
The desired effect of positive steering should be to decrease $p_\text{bt}$, since we aim to push the model towards commitment; conversely, negative steering should increase $p_\text{bt}$. 

Since we study on answer proposals that lead to either commitment or backtracking,
we construct a linear direction for answer commitment from each dataset by taking the DiM between final proposal markers and mid-trace proposal markers. We denote this commitment direction as $v_\text{commit}$ and use it as our baseline.
Given the difference in the frequency of backtracking words that follow a mid-trace marker and a final marker, $v_\text{commit}$ should perform well on steering $p_\text{bt}$.
With $v_\text{commit}$ and $v_\text{conf}$, we apply single token steering at proposal markers with both positive and negative steering.
For each dataset's confidence direction, we select the best steering site and $\alpha$ based on its combined performance on positive and negative steering on the validation split (criteria in App. \ref{app:st_steer_details}). Steering $p_\text{bt}$ is generally monotonic with respect to $\alpha$, so the best $|\alpha|=1$, and the best layer is consistent within a narrow band of layers across all datasets. 

\paragraph{Cross-task generalization of $v_\text{conf}$}
The top row of \autoref{fig:st_steer} plots the validation sweeps for steering $p_\text{bt}$, averaged across each dataset's direction and validation split.
The effective steering regions of each direction are distinct. $v_\text{conf}$'s $\Delta p_\text{bt}$ forms a peak in the middle layers, whereas $v_\text{commit}$'s rises in the middle and stays relatively constant through the final layers.
$v_\text{conf}$ peaks in performance earlier than $v_\text{commit}$ too. 
Interestingly, the peak in steering $p_\text{bt}$ is synchronized with the initial increase in performance for steering VC, suggesting that the layers that begin to separate high and low verbalized confidence markers also carry a transferable signal for controlling $p_\text{bt}$.
Importantly, despite not being fit at the marker position, the peak $\Delta p_\text{bt}$ of $v_\text{conf}$ is comparable to that of $v_\text{commit}$, which \textit{was} fit at the marker position, thus demonstrating cross-task generalization.

We construct a test set of 500 markers by pooling 100 markers from each of the five dataset's test splits. We sample the markers to have an even number of final and mid-trace proposals. 
We evaluate each dataset's direction on the test set and show the results for each model's best $v_\text{conf}$ and $v_\text{commit}$ in \autoref{tab:st_steer_best_small}.
Details on datasets are in App. \ref{app:dataset_rollout_stats} and full results are in \autoref{tab:steer-test-by-dataset} in App. \ref{app:st_steer_details}.
Regarding steering $p_\text{bt}$, as shown in \autoref{tab:steer-test-by-dataset}, each model's best dataset's $v_\text{conf}$ wins on nearly all datasets from the test set. Even though it was fit on only one dataset, it beats out other datasets' $v_\text{conf}$ on their own dataset's test markers.
This suggests that the confidence direction is generalizable rather than dataset-dependent, though some datasets can elicit the extraction of the confidence direction better than others. In particular, the best confidence directions all come from math datasets.
\begin{table}[t]
\centering
\scriptsize
\caption{$p_\text{bt}$ and VC steering with each model's best $v_\text{commit}$ and $v_\text{conf}$ on the marker test set. 
Gap is the absolute distance between the positive- and negative-steering effects.}
\setlength{\tabcolsep}{4pt}
\begin{tabular}{llllrrrllrrr}
\toprule
 &  & \multicolumn{5}{c}{$\Delta p_\text{bt}$} & \multicolumn{5}{c}{$\Delta$ VC} \\
\cmidrule(lr){3-7} \cmidrule(lr){8-12}
Model & Vector & Dataset & Site & Negative & Positive & Gap & Dataset & Site & Negative & Positive & Gap \\
\midrule
\multirow{2}{*}{Olmo 3 32B} & $v_\text{commit}$ & GPQA-D & 43.attn & +0.258 & -0.298 & 0.556 & GSM8K & 24.attn & -0.471 & +0.609 & 1.080 \\
 & $v_\text{conf}$ & AIME 2026 & 23.attn & +0.361 & -0.224 & 0.585 & GPQA-D & 23.attn & -3.502 & +1.215 & 4.717 \\
\midrule
\multirow{2}{*}{Qwen 3 32B} & $v_\text{commit}$ & GPQA-D & 51.mlp & +0.399 & -0.242 & 0.641 & GSM8K & 48.attn & -0.627 & +0.595 & 1.222 \\
 & $v_\text{conf}$ & AIME 2026 & 47.attn & +0.423 & -0.222 & 0.645 & MMLU-Pro & 51.attn & -3.284 & +1.002 & 4.286 \\
\midrule
\multirow{2}{*}{Distill 32B} & $v_\text{commit}$ & MMLU-Pro & 46.attn & +0.432 & -0.382 & 0.814 & AIME 2026 & 48.mlp & -0.290 & +0.273 & 0.563 \\
 & $v_\text{conf}$ & GSM8K & 46.attn & +0.403 & -0.341 & 0.744 & AIME 2026 & 51.attn & -5.198 & +0.891 & 6.089 \\
\bottomrule
\end{tabular}
\label{tab:st_steer_best_small}
\end{table}

\paragraph{Cross-task generalization is asymmetric} 
Since $v_\text{conf}$ transfers to steering $p_\text{bt}$, we check if $v_\text{commit}$ performs well on steering confidence.
As shown in \autoref{fig:st_steer}, this cross-task generalization does not hold for the commitment direction.
In other words, model pathways to emit a backtracking token can be modulated using $v_\text{conf}$ but the model's verbalized confidence cannot be effectively steered using $v_\text{commit}$. 
\revise{\footnote{
\revise{For steering VC, the results in \autoref{tab:st_steer_best_small} show that $v_\text{commit}$ exhibits stronger positive steering than negative steering relative to $v_\text{conf}$, with its positive steering scores being 31-59\% that of $v_\text{conf}$'s. This can be explained as an artifact of the verbalized confidence range having a max value of 9. If an answer proposal marker has a high initial confidence, i.e. 8, positive steering saturates with $\Delta VC = +1$, thus artifically deflating the capability of $v_\text{conf}$ in the positive confidence steering setting. We explore this in App. \ref{app:st_steer_details:vc_steer_anal}.}
}}
The two directions are numerically different too: within the 10-layer window centered at $v_\text{conf}$'s best site for steering $p_\text{bt}$, the cosine similarity between $v_\text{conf}$ and $v_\text{commit}$ is 0.028 on average and at most 0.122 across all models.
We also 
demonstrate in App. \ref{app:ablation_misc} that the confidence direction cannot be trivially explained as merely a next-token probability decoder
or a position detector.

\subsection{Direction Analysis}
\paragraph{Confidence direction decodes meaningfully} 
\label{sec:4logit-lens}
We take the logit lens  \citep{nostalgebraist2020logitlens} of the best $v_\text{conf}$ per model and find that the top tokens correspond to concepts semantically related to confidence.
Logit lens with Olmo 3 32B $v_\text{conf}$ decodes tokens like ``questionable" and ``controversial" in the negative direction, and ``flawless" and ``perfected" in the positive direction. Qwen 3 32B and Distill 32B decode similar concepts in a multilingual space.
\revise{
On the other hand, the logit lens of $v_\text{commit}$ operates at the syntactic level. The negative direction decodes backtracking tokens like ``However'' and ``But'' in English or Chinese. 
For Olmo 3 32B, the negative direction also decodes affirmatory adverbs like ``Therefore'' and ``Thus'', and its positive direction's top token is \texttt{</}, the first token of \texttt{</think>}. 
This suggests that steering with Olmo 3 32B's $v_\text{commit}$ directly promotes the reasoning termination tag and suppresses adverbs leading to additional reasoning,
which is consistent with the construction of $v_\text{commit}$ from final and mid-trace proposal markers.
The positive directions of Qwen 3 32B and Distill 32B are less interpretable.
}
Unfiltered top tokens are in \autoref{tab:logit_lens_vconf} and \autoref{tab:logit_lens_vcommit}.

\revise{
\paragraph{Projection correlation}
\label{sec:4correlations}
To test whether latent confidence tracks commitment without intervention, we project proposal marker activations onto
$v_\text{commit}$ or $v_\text{conf}$ and compute Spearman correlations with answer commitment on the test set (\autoref{tab:marker-corr-commit}).
As a reference, $v_\text{commit}$, which was directly fit on commitment labels, has $\rho \approx 0.80$ ($\rho_{\max}=0.87$).
Olmo 3 32B and Qwen 3 32B, projections onto $v_\text{conf}$ have $\rho=0.49,0.64$ respectively, which exceed that of verbalized confidence on the test set (0.33, 0.26). 
This disparity is largest on incorrect rollouts, where verbalized confidence is uncorrelated ($\rho = 0.09,-0.03)$ while $v_\text{conf}$ projection remains at $\rho = 0.36, 0.58$. 
On Distill 32B, the projection has no correlation with answer commitment ($\rho=-0.00$), yet projections from all three models do correlate with $p_\text{bt}$ ($\rho \in [-0.68, -0.46]$; \autoref{tab:marker-corr-pback}).
We discuss further in \cref{sec:conclusion}.
}

\section{Multi-Token Steering}
\label{sec:5}
Since steering $p_\text{bt}$ is a limited approximation of answer commitment behavior, we expand our evaluation to more realistic multiple token generation settings.
Specifically, we compare our activation-based steering with linguistic interventions (\cref{sec:5cot_vs_steer}) and evaluate multi-token steering from an answer proposal (\cref{sec:5postmarker}) and from scratch (\cref{sec:5steer_scratch}). Unless specified otherwise, experiments use the best confidence and commitment directions based on steering $p_\text{bt}$ in \autoref{tab:st_steer_best_small}.


\subsection{Surface-level confidence versus latent confidence}
\label{sec:5cot_vs_steer}
Models may naturally vary their linguistic confidence or propensity to commit to an answer over the course of the reasoning trace. For example, confident proposals such as \texttt{\small Therefore, I am confident that the answer is X} are more likely to appear at the end of the trace than the middle. Similarly, models may produce consecutive answer proposal sentences when they are on the verge of committing. Thus, is a confidence signal in the surface-level CoT or the model's hidden states more effective at causing the model to commit to its answer?

We filter the 500 marker test set for mid-trace markers whose sampled next token is a backtracking marker.
At these mid-trace proposals, we either replace the proposal sentence with a confident proposal or append the confident proposal.
We then measure the change in the next token probability of backtracking and if the model generates \texttt{</think>} within the next 128 tokens.
We compare this against steering the last token of the answer proposal sentence.
In this manner, we measure how much the answer proposal's surface-level confidence interacts with the internal confidence of the model.
We generate 4 rollouts per marker to get a more accurate measure of $\Delta_{\texttt{</think>}}$.

\begin{table}[t]
  \centering
  \scriptsize
  \setlength{\tabcolsep}{4pt}
  \caption{Intervening on linguistic confidence versus latent confidence. $\Delta_\texttt{</think>}$ and $\Delta p_{\mathrm{bt}}$ are reported as percentages against each model's positive and negative baselines (in \autoref{tab:cot_vs_steer_controls}). $\Delta_{\texttt{think}} > 0$ implies quicker termination of reasoning.}
  \label{tab:cot_vs_steer_primary_append}
  \begin{tabular}{llcccccc||cccccc}
    \toprule
    \multirow{3}{*}{Model} & \multirow{3}{*}{Metric} & \multicolumn{6}{c||}{Positive: confident proposal (mid-trace markers)} & \multicolumn{6}{c}{Negative: hedged proposal (commitment markers)} \\
    \cmidrule(lr){3-8} \cmidrule(lr){9-14}
     &  & \multicolumn{2}{c}{replacement} & \multicolumn{2}{c}{append} & \multicolumn{2}{c||}{steer ($\alpha$)} & \multicolumn{2}{c}{replacement} & \multicolumn{2}{c}{append} & \multicolumn{2}{c}{steer ($\alpha$)} \\
    \cmidrule(lr){3-4} \cmidrule(lr){5-6} \cmidrule(lr){7-8} \cmidrule(lr){9-10} \cmidrule(lr){11-12} \cmidrule(lr){13-14}
     &  & neu. & conf. & neu. & conf. & 0.5 & 1 & neu. & unc. & neu. & unc. & $-$0.5 & $-$1 \\
    \midrule
    \multirow{2}{*}{Olmo 3 32B} & $\Delta_{\texttt{</think>}}$ & $+$1.6 & $+$3.0 & $+$6.2 & \textbf{$+$15.9} & $+$8.2 & \underline{$+$10.5} & $-$4.9 & $-$18.4 & $-$0.8 & \underline{$-$19.0} & $-$12.6 & \textbf{$-$22.1} \\
     & $\Delta p_{\mathrm{bt}}$ & $-$0.9 & $+$1.0 & $+$6.3 & $-$2.7 & \underline{$-$27.2} & \textbf{$-$47.9} & $+$9.6 & $+$30.3 & $+$4.2 & \underline{$+$36.4} & $+$22.8 & \textbf{$+$40.1} \\
    \midrule
    \multirow{2}{*}{Qwen 3 32B} & $\Delta_{\texttt{</think>}}$ & $+$2.9 & $+$6.6 & \underline{$+$18.2} & \textbf{$+$27.3} & $+$11.8 & $+$15.6 & $-$3.1 & $-$15.8 & $+$1.8 & $-$4.6 & \underline{$-$17.1} & \textbf{$-$27.8} \\
     & $\Delta p_{\mathrm{bt}}$ & $+$0.4 & $+$0.9 & $-$3.4 & $-$13.9 & \underline{$-$41.3} & \textbf{$-$70.3} & $+$9.5 & \underline{$+$31.6} & $-$1.6 & $+$11.9 & $+$30.9 & \textbf{$+$55.3} \\
    \midrule
    \multirow{2}{*}{Distill 32B} & $\Delta_{\texttt{</think>}}$ & $+$1.8 & $+$15.2 & $+$23.8 & \textbf{$+$44.2} & $+$21.9 & \underline{$+$26.2} & $-$1.8 & \underline{$-$16.7} & $+$5.1 & $-$3.7 & $-$13.4 & \textbf{$-$21.9} \\
     & $\Delta p_{\mathrm{bt}}$ & $+$6.5 & $-$6.2 & $-$10.8 & $-$31.0 & \underline{$-$46.3} & \textbf{$-$70.8} & $+$9.1 & \underline{$+$32.5} & $-$4.7 & $+$16.0 & $+$29.5 & \textbf{$+$40.6} \\
    \bottomrule
  \end{tabular}
\end{table}

We test the reverse setting as well:
At final proposal markers, we filter for commitment markers that are within 12 tokens from the \texttt{</think>} token.
We then either replace or append that proposal sentence with an uncertain sentence containing the answer value, and measure the increase in next token backtracking probability and the fraction of responses that end reasoning within the next 256 tokens. 
We choose 256 instead of 128 to allow the model some buffer for possible wordy "superficial" backtracking.
We apply the same steering intervention with negative $\alpha$.

We employ three linguistically confident answer proposals and three linguistically uncertain sentences hedging the answer proposal. We also use a neutral sentence to control for the variation resulting from manipulating the CoT out of the model's natural distribution. The sentences are listed in \autoref{tab:cot_vs_steer_sentences}.
Results are shown \autoref{tab:cot_vs_steer_primary_append}, with the results of the three confident/uncertain sentences averaged to a single column. 

Answer proposal replacement has a modest effect that is similar to or slightly worse than steering with $\alpha=0.5$ across both settings.
The effects of answer proposal addition are asymmetric. Although it is most effective at answer commitment in the positive setting, it is unreliable in the negative setting for Qwen 3 32B and Distill 32B, performing worse than proposal replacement.
Steering with $|\alpha| = 1$ outperforms both CoT manipulation methods in the negative setting, and it trails behind answer proposal addition in the positive setting.

Overall, we find that adding latent confidence only to the last token of the answer proposal sentence is more reliable at modulating answer proposal backtracking compared to CoT confidence manipulation.
\stephen{phrase better, mainly want to emphasize that single token steering is a minimal intervention.}


\subsection{Confidence steering controls reasoning length}
\label{sec:5postmarker}
\cref{sec:4st_steer} and \cref{sec:5cot_vs_steer} demonstrate that the confidence direction controls $p_\text{bt}$ and $\Delta_\texttt{</think>}$ through single token steering, but how does a consistent multi-token steering presence affect the rest of the model's generations? To answer this question, we continue generation from an answer proposal while steering at each generated token to measure how much a continued steering presence on newly generated tokens affects the model's behavior toward answer commitment and backtracking.

We continue steering the model after the last answer proposal token to measure if the direction reduces the number of answer proposals while maintaining accuracy.
We sample \revise{four continuations} after each proposal with temperature=0.6, top-p=0.95, and max. response length of 32,768 tokens, and sweep across alpha values (0.1, 0.25, 0.5, 0.75) and (-0.1, -0.25, -0.5). 
\revise{
We also sample four unsteered $\alpha=0$ continuations as a control.
For large $\alpha$ values that result in degenerate completions, we sample only 1 continuation, as these CoTs typically hit the response limit and provide no answer.}

\begin{figure}[t]
    \centering
    \includegraphics[width=0.99\linewidth]{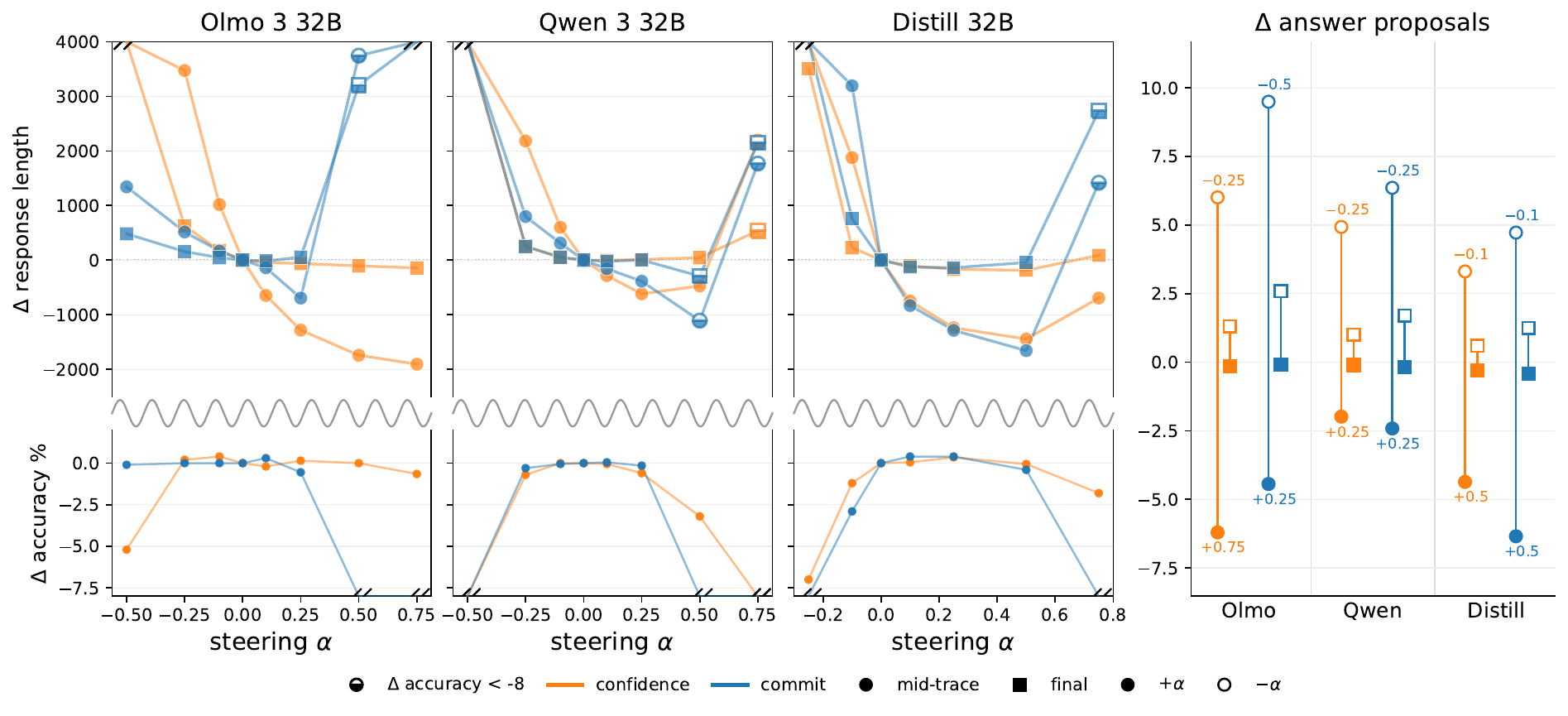}
    \caption{Post Marker Continuation Steering Length and accuracy shift vs $\alpha$. With $\alpha>0$, negative $\Delta$ response length is better. With $\alpha<0$, positive $\Delta$ response length is better. Confidence steering matches or outperforms steering with the commitment direction, when constrained to keep $\Delta$ accuracy close to 0. Each model's best $\alpha$'s are indicated on the right. Raw results in \autoref{tab:post-marker-steer-compact}. }
    \label{fig:post-marker-steer}
\end{figure}

We measure the change in the remaining response length, overall accuracy, and, for each model and vector's best positive and negative $\alpha$, the number of additional answer proposals.
We report separate results for mid-trace and final markers since final proposals are already near reasoning termination. 
As shown in \autoref{fig:post-marker-steer}, $v_\text{conf}$ changes answer commitment behavior in both directions, 
comparable to $v_\text{commit}$.
Using best $\alpha$'s, positive steering with $v_\text{conf}$ decreases average response length by $1323$ and $107$ tokens on mid-trace and final markers respectively at little cost to accuracy. Negative steering with $v_\text{conf}$ increases response length by $2511$ and $370$ tokens on mid-trace and final markers.
On mid-trace proposals, positive steering with $v_\text{conf}$ reduces answer proposal count by \revise{6.21, 1.98, and 4.36} on Olmo 3 32B, Qwen 3 32B, and Distill 32B, respectively. Negative steering conversely increases answer proposal count by \revise{6.01, 4.93, and 3.31}, respectively.
Steering final proposal markers with $+\alpha$ slightly increases response length on Qwen 3 32B, though this is largely sampling noise, since the model is already close to ending generation.
\revise{Additional analysis on rollouts with more than one unique answer are in App. \ref{app:postmarkersteer}.}


\subsection{Steering at Inference Time}
\label{sec:5steer_scratch}
Steering the model from an answer proposal marker can reliably modulate response length and the number of answer proposals.
In particular, positive steering after the answer proposal position with the latent confidence direction can reduce token count. However, answer proposal positions are unknown at inference time, so we test positive confidence steering in the inference setting too.

\begin{table}[t]
    \centering
    \scriptsize
    \caption{Confidence steering at inference time, with token delay $d$. Steering with $d=0$ decreases response length by 8.4-20.0\% across models but slightly degrades accuracy. Delaying steering by $d=5000$ tokens helps recover the accuracy loss. \revise{Confidence intervals are in \autoref{tab:steer-scratch-cis}.}}
    \setlength{\tabcolsep}{4pt}
\begin{tabular}{ll|rr|rr|rr|rr|rr|rr}
\toprule
Model & Arm & \multicolumn{2}{c}{AIME 2026} & \multicolumn{2}{c}{GSM8K} & \multicolumn{2}{c}{OMEGA-500} & \multicolumn{2}{c}{GPQA-D} & \multicolumn{2}{c}{MMLU-Pro} & \multicolumn{2}{c}{Avg $\Delta$} \\
\cmidrule(lr){3-4}\cmidrule(lr){5-6}\cmidrule(lr){7-8}\cmidrule(lr){9-10}\cmidrule(lr){11-12}\cmidrule(lr){13-14}
 & & Acc & Len & Acc & Len & Acc & Len & Acc & Len & Acc & Len & $\Delta$Acc & $\Delta$Len \\
\midrule
\multirow{3}{*}{Olmo 3 32B} & Unsteered & 80.8 & 15788 & 95.4 & 1939 & 61.9 & 16237 & 61.5 & 12686 & 71.9 & 6086 & -- & -- \\
 & $\alpha$=0.1, $d$=0 & 75.8 & 14328 & 94.8 & 1512 & 59.2 & 14399 & 58.8 & 10966 & 71.1 & 4864 & -2.3 & -15.2\% \\
 & $\alpha$=0.1, $d$=5000 & 79.2 & 14850 & 95.2 & 1912 & 59.9 & 14835 & 62.1 & 11357 & 71.4 & 5547 & -0.7 & -7.1\% \\
\midrule
\multirow{3}{*}{Qwen 3 32B} & Unsteered & 71.7 & 14537 & 94.5 & 1820 & 53.5 & 14857 & 67.0 & 7781 & 76.0 & 3904 & -- & -- \\
 & $\alpha$=0.1 $d$=0 & 70.0 & 13580 & 93.8 & 1494 & 54.4 & 14070 & 67.4 & 7299 & 75.0 & 3660 & -0.4 & -8.4\% \\
 & $\alpha$=0.1, $d$=5000 & 70.8 & 13428 & 94.4 & 1844 & 55.4 & 13917 & 67.7 & 7713 & 76.2 & 3850 & +0.4 & -3.0\% \\
\midrule
\multirow{3}{*}{Distill 32B} & Unsteered & 63.3 & 12088 & 94.6 & 1502 & 45.6 & 12878 & 60.5 & 6753 & 70.2 & 3717 & -- & -- \\
 & $\alpha$=0.1 $d$=0 & 60.8 & 9846 & 95.0 & 1112 & 40.8 & 10533 & 59.2 & 5471 & 72.5 & 3028 & -1.2 & -20.0\% \\
 & $\alpha$=0.1, $d$=5000 & 58.3 & 10831 & 94.9 & 1372 & 44.8 & 11030 & 62.2 & 6145 & 70.6 & 3368 & -0.7 & -10.4\% \\
\bottomrule
\end{tabular}
    \label{tab:steer-scratch}
\end{table}

We steer with a small $\alpha=0.1$ to avoid degenerate rollouts.
Since reasoning models typically spend the initial part of the reasoning trace exploring and formulating plans, with answer proposals arising later into the trace, we test a second setting where we delay steering until $d=5000$ tokens have been generated to allow the model to first reason without intervention.

Results are shown in \autoref{tab:steer-scratch}. Without a delay, steering with $\alpha=0.1$ reduces response length by 8.4-20.0\% across models 
\revise{
but also degrades accuracy on more difficult datasets. For example, Olmo 3 32B's AIME accuracy drops from 80.8\% to 75.8\%.
This likely arises because the model's initial reasoning steps strongly influence later reasoning steps, so increasing confidence too early causes the model to be overconfident on pivotal steps where backtracking is useful.
Applying $d=5000$ delay recovers most of the accuracy drop but also reduces the gains in reasoning efficiency, demonstrating an accuracy-efficiency tradeoff on harder datasets}. 
As we do not optimize for the best $\alpha$ or use custom delays for various datasets, we present these results not as a practical efficiency method but as a proof of concept that the effect of confidence steering holds without knowing the exact positions of the proposals in advance.
\stephen{maybe cite some stuff on efficient reasoning?}
We use vLLM \citep{kwon2023efficientmemorymanagementlarge} for all rollouts, and steering has negligible impact on inference speed.


\begin{table}[htbp]
  \centering
  \footnotesize
  \caption{Effect of regenerating an answer proposal with $\alpha=+0.5$ confidence steering on Olmo 3 32B (AIME 2026, problem index 27). Both continuations share the same prefix. Adding confidence causes the model to consolidate and commit. }
  \label{tab:steering_qualitative_olmo3_32b}
  \begin{tabular}{@{}>{\raggedright\arraybackslash}p{0.47\linewidth}>{\raggedright\arraybackslash}p{0.47\linewidth}@{}}
    \toprule
    \multicolumn{2}{@{}>{\raggedright\arraybackslash}p{0.96\linewidth}@{}}{%
      \textbf{Problem.} Call finite sets of integers $S$ and $T$ \emph{cousins} if
      (i) $S$ and $T$ have the same number of elements,
      (ii) $S$ and $T$ are disjoint, and
      (iii) the elements of $S$ can be paired with the elements of $T$ so that
      the elements in each pair differ by exactly $1$.
      For example, $\{1,2,5\}$ and $\{0,3,4\}$ are cousins.
      Suppose that the set $S$ has exactly $4040$ cousins.
      Find the least number of elements the set $S$ can have. \textbf{Ground Truth.} 107.} \\
    \midrule
    \multicolumn{2}{@{}>{\raggedright\arraybackslash}p{0.96\linewidth}@{}}{%
      \textbf{Shared prefix (through marker 29).}
      \itshape Thus $S=\{0,2,3\}$ has only 1 cousin. Not helpful. This isn't better.} \\
    \midrule
    \textbf{Original (Unsteered)} & \textbf{Regenerated (Steered $\alpha=0.5$)} \\
    \midrule
    \itshape Given that I can't find a better configuration, I think the answer is 107. But since this is an Olympiad problem, maybe the answer is smaller. Perhaps my initial assumption about\dots
    &
    \itshape After exploring various configurations and consistently arriving at 107 as the minimal, I think that's the answer. So I'll go with $\boxed{107}$ \newline\newline
    **Final Answer** \newline $\boxed{107}$ \newline
    \texttt{</think>} \newline
    To solve the problem\dots \\
    \midrule
    Proposes 107, then immediately backtracks.
    & Proposes 107, affirms it, and exits the reasoning trace. \\
    \bottomrule
  \end{tabular}
\end{table}
\paragraph{Steering answer proposal linguistics}
\label{sec:5steer_qualitative}
Lastly, having established that the latent confidence direction modulates trace-level statistics of response length and answer proposal count, we qualitatively inspect whether steering with the latent confidence direction affects the linguistic nature of the answer proposal sentence itself. 
At answer proposal markers, we regenerate the answer proposal sentence with positive and negative confidence steering at $\alpha=\pm0.5$ and continue steering up to 128 tokens.
With positive steering, models typically regenerate their original proposal, or reword more confidently (see \autoref{tab:steering_qualitative_olmo3_32b}).
Similarly, with negative steering, regenerated answer proposals express uncertainty (\autoref{tab:steering_qualitative_r1qwen_32b}) or abandon the proposal entirely in favor of backtracking immediately (\autoref{tab:steering_qualitative_qwen3_32b}).

\section{Conclusion}
\label{sec:conclusion}
To better understand the relationship between model confidence and backtracking, we provide evidence that latent confidence signals can be extracted through verbalized confidence elicitation and accumulated to approximate a linear confidence direction. 
This confidence direction robustly transfers to steering next token probability of backtracking, number of additional answer proposals, and reasoning length, demonstrating its ability to control answer commitment behavior. 
Interestingly, this cross-task generalization is asymmetric, as a linear direction fit to separate answer commitment and backtracking can steer answer commitment but cannot steer verbalized confidence.
We also provide comparisons between and interactions with surface-level confidence and latent confidence, finding that 1) single token steering elicits more reliable effects than increasing the linguistic confidence of the answer proposal, and 2) the effects of confidence steering are reflected in the model's linguistic confidence.

\paragraph{Limitations and Future Work}
Since model internal confidence is likely a multi-faceted representation that cannot be captured linearly, we do not claim that model confidence is fully explained by our methodology.
Rather, we defend that the linear direction separating the model's verbalized assessment of its confidence transfers effectively towards steering behavior at answer proposals.
We also notice that unsteered reasoning traces occasionally terminate with a resignation phrase like \texttt{\small Given how much time I have spent, I think I will go with X}. In these cases, the model is not verbally confident, but it commits anyway, demonstrating that high confidence may be a sufficient but not necessary criterion for answer commitment. 
\revise{
Additionally, the projection onto latent confidence correlates with answer commitment in Olmo 3 32B and Qwen 3 32B, which undergo RLVR, but not in Distill 32B, although it correlates with backtracking token probability in all three models (\cref{sec:4correlations}).
}
A natural next step is to measure how the linear direction evolves across reasoning post-training to test whether RL strengthens the link to answer commitment.
\revise{
Finally, steering shows that the latent confidence direction can modulate backtracking, but this does not guarantee that models naturally use this direction to the extent that steering applies it, which fits into the broader interpretability research question on the ``naturalness'' vs ``causality'' of steering \citep{mishra2026steeredllmactivationsnonsurjective}, and which we leave to future work.
}

\subsection*{AI use statement}

In this work, we used generative AI tools to design or provide feedback on research methodology or experiments, implement methods, clean and reformat datasets, support qualitative and thematic data analysis, and interpret results.
We have not used generative AI tools to generate synthetic data sets, help develop theoretical models or conceptual frameworks, formulate mathematical claims, provide critical ingredients for proving mathematical claims, assist in the writing of proofs, propose or refine hypotheses, or assist with translation.
Additionally, we used generative AI tools to assist with plot and figure generation and to identify relevant related work. The rest of the disclosure tasks are not applicable to this work.
We have reviewed all AI-assisted work.
LLM-generated code was verified and tested for correctness.
We take responsibility for the final content of this work,
including text, claims or artifacts produced with the aid of generative AI.





\subsection*{Reproducibility statement}
All models and datasets used in this work are publicly available, and we
report the sampling parameters (temperature, top-$p$, maximum response
length) and prompt subsampling used to generate rollouts in
\cref{sec:3answerproposals}. The LLM judge pipeline for identifying answer proposals is
described in App. \ref{app:judge}, including the verbatim prompt templates for
both stages (\autoref{tab:judge-prompt-math}, \autoref{tab:judge-prompt-verify}),
the manual review procedure, and
agreement against author annotations (\autoref{tab:human_llm_agreement}).
\cref{sec:3dimconstruction} specifies the verbalized confidence prompt, the
Difference-in-Means construction, and the train/validation/test split
procedure. Per-model and per-dataset marker counts are in
\autoref{tab:markers_per_dataset} and \autoref{tab:test_marker_splits}. 
App. \ref{app:st_steer_details} gives the steering
implementation, the validation sweep, and the exact best-site selection criterion. We use
vLLM for all steered rollouts. We intend to release code for direction
extraction, steering, and evaluation upon publication.




\bibliography{iclr2027_conference}
\bibliographystyle{iclr2027_conference}

\appendix
\section{LLM Judge Pipeline}
\label{app:judge}

\paragraph{Motivation for Judge Pipeline}
Extracting answer proposals from a reasoning trace requires scanning rollouts that are thousands of tokens long, which is impractical for humans to do.
Furthermore, relying solely on regex based on the model's final prediction would ignore any intermediate answer values the model had proposed and later revised. 
A similarly simple option is to find all sentences in the trace with the word ``answer". However, answer proposal sentences can vary in lexical style, and a sentence does not necessarily need to have the word ``answer'' for the model to commit at that sentence.
An LLM judge can handle the majority of the labeling at scale, with human annotation as a final pass at the end. 
Since we aim to study representations at answer proposal markers, the LLM judge pipeline should have high precision. In other words, for any answer sentence the LLM judge selects to be an answer proposal, the human should have also selected it to be an answer proposal.
Additionally, high recall on the final answer proposal is critical, since only one proposal per rollout is the final answer proposal, and it is used to create the commit class for our Difference-in-Means vector.
High recall for mid-trace answer proposals is a lower priority, as there are typically many mid-trace answer proposals within a reasoning trace and only one final answer proposal.

We use a two step judge pipeline:
\begin{enumerate}
    \item For each rollout, we ask the judge to return all unique answer values proposed by the model, including proposed values that were later rejected or revised. We skip rollouts that do not have a final prediction.
    \item For each rollout, we tag each sentence in the reasoning trace with a [[Sk]] marker if it contains any answer value returned from step 1, where k is the index id that distinguishes the candidate markers. This filters down the number of sentences the judge has to evaluate. Then, we pass the whole rollout to the judge and ask the judge to select the markers corresponding to the answer proposal sentences.
\end{enumerate}

For MCQA, we skip step 1 and instead use regex to tag all sentences that contain a standalone capital letter from ``A"-``D" for GPQA-D and ``A"-``J" for MMLU-Pro. The regex rule ensures that the letter does not get selected if it has adjacent letters, digits, or the punctuation [-, ', \textunderscore, \textasciicircum , \textbackslash, =]. Thus (C), C. and option C do match. For letters ``A" and ``I", which are English words, if the letter is followed by a space and a lowercase word (or, for ``A", a number), it is treated as the article or pronoun and skipped (``A benzene ring", ``I think"). If an option cue comes before it, such as ``option", ``choice", ``answer" or ``letter", we do select it.

\autoref{tab:judge-prompt-math} shows the prompt for step 1. \autoref{tab:judge-prompt-verify} and \autoref{tab:judge-prompt-verify-mcq} show the prompts for step 2 for math and MCQA, respectively. Finally, after the judge pipeline, we manually check each rollouts whose last candidate marker was not selected by the judge, and select the candidate marker if it is indeed an answer proposal.
If the last sentence is a hedging statement, such as "So, maybe the correct answer is not listed, but among the options, option (A) is the closest." we do select it, because the model commits with this proposal. 
If two consecutive sentences both arrive at an answer to the question, i.e. ``So the answer is 10. Therefore, I believe the answer is 10", we treat both sentences as answer proposals. This is because it is plausible that the model commits after generating the first answer proposal sentence. Typically, this pattern occurs near the reasoning termination.

\paragraph{Manual check against golden labels}
We manually annotate answer proposals on 30 rollouts from Olmo 3 32B on AIME 2026, for a total of 332 selected markers, and 30 rollouts from Olmo 3 32B on GPQA-D, for a total of 209 selected markers.
We treat these as golden labels, and use them to iteratively design the judge prompt templates for math questions and MCQA questions for all models. Precision and Recall of the final judge pipeline (without the manual check) are shown in \autoref{tab:human_llm_agreement}.

\begin{table}[htbp]
    \centering
    \small
    \begin{tabular}{l|c|c}
        Question Format Type & Precision & Recall \\
        \toprule
        Math & 0.9129 & 0.7892 \\
        \midrule
        MCQA & 0.8676 & 0.9091 \\
        \bottomrule
    \end{tabular}
    \caption{Human-LLM Agreement for our Judge Pipeline.}
    \label{tab:human_llm_agreement}
\end{table}

\begin{table}[htbp]
  \centering
  \caption{Prompt given to the judge model to extract the set of unique proposed
  answers from a reasoning trace on math tasks. \texttt{\{question\}} and
  \texttt{\{response\}} are replaced with the problem statement and the model's
  full rollout.}
  \label{tab:judge-prompt-math}
  \begin{tabular}{@{}>{\ttfamily\scriptsize\raggedright\arraybackslash}p{0.95\linewidth}@{}}
    \toprule
    You are an expert at parsing language model thinking traces. \\
    \addlinespace
    Given a question and a language model's response, return a list of all unique
    solutions proposed throughout the response. Do not treat any intermediate
    values as solutions. If the same answer is formatted in multiple ways, then
    include each format in the list. Copy each answer verbatim. Below are examples
    of answers formatted in multiple ways: \\
    \addlinespace
    \hspace{1em}- "1/2", "\textbackslash{}frac\{1\}\{2\}", "\textbackslash{}dfrac\{1\}\{2\}" \newline
    \hspace{1em}- "5" and "\textbackslash{}boxed\{5\}" \newline
    \hspace{1em}- "6+9i" and "6 + 9i" \\
    \addlinespace
    Here is the question: \\
    \addlinespace
    <question> \newline
    \{question\} \newline
    </question> \\
    \addlinespace
    Here is the response: \\
    \addlinespace
    <response> \newline
    \{response\} \newline
    </response> \\
    \addlinespace
    Do not solve the problem yourself. List all unique solutions proposed
    throughout the response. Do not treat any intermediate values as a solution.
    For example, if the question asks for a fraction expressed as the sum of the
    numerator and denominator, do not treat the fraction as an answer. If an
    answer is formatted in multiple ways, then include each format in the list.
    Copy each answer verbatim. \\
    \addlinespace
    Return the list of answers in a JSON:
    \{"answers": [<str>, <str>, ...]\}. \\
    \bottomrule
  \end{tabular}
\end{table}

\begin{table}[htbp]
  \centering
  \caption{Prompt given to the judge model to select, from a set of inserted
  sentence markers, those at which the model commits to an answer.
  \texttt{\{question\}} and \texttt{\{response\}} are replaced with the problem
  statement and the marker-annotated rollout.}
  \label{tab:judge-prompt-verify}
  \begin{tabular}{@{}>{\ttfamily\scriptsize\raggedright\arraybackslash}p{0.95\linewidth}@{}}
    \toprule
    You are an expert at parsing language model thinking traces. \\
    \addlinespace
    You will be shown a question and a language model's response. A tag
    {}[[Sk]] has been inserted at the END of certain sentences, where k is the
    marker id. The tags are annotations and were not produced by the model. \\
    \addlinespace
    For each tagged marker, decide whether the sentence BEFORE the marker is one
    in which the model commits to an answer to the prompt. Select only those
    markers. Do NOT select a marker whose preceding sentence is an intermediate
    step or does not settle on an answer to the question. \\
    \addlinespace
    Here is the question: \\
    \addlinespace
    <prompt> \newline
    \{question\} \newline
    </prompt> \\
    \addlinespace
    Here is the model's response: \\
    \addlinespace
    <response> \newline
    \{response\} \newline
    </response> \\
    \addlinespace
    A tag [[Sk]] has been inserted at the END of certain sentences, where k is
    the marker id. The tags are annotations and were not produced by the model. \\
    \addlinespace
    First, identify precisely which quantity the problem asks the language model
    to find. For each tagged marker, decide whether the sentence BEFORE the
    marker is one in which the model commits to an answer for that exact
    quantity. Select only those markers. \\
    \addlinespace
    Do NOT select a marker if: \\
    \addlinespace
    \hspace{1em}- the preceding sentence is an intermediate step or auxiliary
    quantity that does not settle on an answer to the question. \\
    \hspace{1em}- the preceding sentence ends with a question mark or has an
    explicitly unconfident clause, such as "but let me make sure". It is fine if
    the sentence AFTER the tag is uncertain. \\
    \hspace{1em}- the preceding sentence contains no English words - it is only
    symbols, numbers and operators. "180 + 24 = 204." and "m+n=100+13=113." are
    NOT proposals. A brief computation phrased as the answer, such as "100 and 13
    are coprime, so m=100, n=13, so m+n=113 is the answer" DOES count. \\
    \hspace{1em}- the value is only hedged as a possibility, such as "maybe the
    answer is 5" or "perhaps it's 5" - UNLESS the sentence also decides to go
    with it, as in "Maybe the answer is 113, so I'll go with that", which DOES
    count. A neutral hypothetical such as "the total would be \$90" DOES count. \\
    \addlinespace
    If you are unsure about a marker, it is better not select it. The final
    tagged markers are the most important to evaluate. For each selected marker,
    record its id and the model's answer. \\
    \addlinespace
    Return JSON: \{"answers": [\{"marker\_id": <int>, "answer": <string>\},
    ...]\}. If no markers qualify, return an empty list. \\
    \bottomrule
  \end{tabular}
\end{table}

\begin{table}[htbp]
  \centering
  \caption{Prompt given to the judge model to select, from a set of inserted
  sentence markers, those at which the model commits to an answer choice on
  multiple-choice tasks. \texttt{\{question\}} and \texttt{\{response\}} are
  replaced with the problem statement and the marker-annotated rollout.}
  \label{tab:judge-prompt-verify-mcq}
  \begin{tabular}{@{}>{\ttfamily\scriptsize\raggedright\arraybackslash}p{0.95\linewidth}@{}}
    \toprule
    You are an expert at parsing language model thinking traces. \\
    \addlinespace
    You will be shown a multiple choice question and a language model's response.
    A tag {}[[Sk]] has been inserted at the END of certain sentences, where k is
    the marker id. The tags are annotations and were not produced by the model. \\
    \addlinespace
    For each tagged marker, decide whether the sentence BEFORE the marker is one
    in which the model commits to an answer choice to the question. Select only
    those markers. Do NOT select a marker whose preceding sentence is an
    intermediate step or does not settle on an answer to the question. \\
    \addlinespace
    Here is the question: \\
    \addlinespace
    <prompt> \newline
    \{question\} \newline
    </prompt> \\
    \addlinespace
    Here is the model's response: \\
    \addlinespace
    <response> \newline
    \{response\} \newline
    </response> \\
    \addlinespace
    A tag [[Sk]] has been inserted at the END of certain sentences, where k is
    the marker id. The tags are annotations and were not produced by the model. \\
    \addlinespace
    First, identify precisely what the problem asks the language model to find.
    For each tagged marker, decide whether the sentence BEFORE the marker is one
    in which the model commits to an answer choice. Select only those markers. \\
    \addlinespace
    Do NOT select a marker if: \\
    \addlinespace
    \hspace{1em}- the preceding sentence is an intermediate step or auxiliary
    quantity that does not settle on an answer to the question. \\
    \hspace{1em}- the preceding sentence ends with a question mark or has an
    explicitly unconfident clause, such as "but let me make sure". It is fine if
    the sentence AFTER the tag is uncertain. \\
    \hspace{1em}- the preceding sentence does not contain an answer choice. \\
    \hspace{1em}- the value is only hedged as a possibility, such as "maybe the
    answer is C" or "perhaps it's C" - UNLESS the sentence also decides to go
    with it, as in "Maybe the answer is C, so I'll go with that", which DOES
    count. A neutral hypothetical such as "the total would be A" DOES count. \\
    \addlinespace
    If you are unsure about a marker, it is better not select it. The final
    tagged markers are the most important to evaluate. For each selected marker,
    record its id and the model's answer. \\
    \addlinespace
    Return JSON: \{"answers": [\{"marker\_id": <int>, "answer": <string>\},
    ...]\}. If no markers qualify, return an empty list. \\
    \bottomrule
  \end{tabular}
\end{table}

\FloatBarrier

\section{Dataset Details}
\label{app:dataset_rollout_stats}

\paragraph{Marker Splits}
\autoref{tab:markers_per_dataset} shows the total number of rollouts and markers for each model and dataset, which is used to form the train, validation, and test splits to construct the confidence direction and commitment direction.
\autoref{tab:test_marker_splits} details the marker test sets for each model. 100 markers from each dataset's test split are sampled to form the test set.

\paragraph {Rollout Statistics}
\autoref{tab:backtrack-freq} shows backtracking frequency per proposal marker type on each model and dataset.
\autoref{tab:unique-vs-proposed} shows unique answers vs.\ answer proposals per rollout per model and dataset.

\begin{table}[htbp]
    \centering
    \scriptsize
    \caption{Marker Splits for each Dataset and Model after passing the model through the judge pipeline. The number in parentheses indicates the number of prompts used to generate rollouts for each dataset. We subsample larger datasets randomly to have 200 prompts. }
    \label{tab:markers_per_dataset}
    \setlength{\tabcolsep}{3pt}
\begin{tabular}{lrrrrrrrrrr}
\toprule
 & \multicolumn{2}{c}{AIME 2026 (30)} & \multicolumn{2}{c}{GSM8K (200)} & \multicolumn{2}{c}{OMEGA-500 (200)} & \multicolumn{2}{c}{GPQA-D (198)} & \multicolumn{2}{c}{MMLU-Pro (200)} \\
\cmidrule(lr){2-3} \cmidrule(lr){4-5} \cmidrule(lr){6-7} \cmidrule(lr){8-9} \cmidrule(lr){10-11}
Model & \# Rollouts & \# Markers & \# Rollouts & \# Markers & \# Rollouts & \# Markers & \# Rollouts & \# Markers & \# Rollouts & \# Markers \\
\midrule
Olmo 3 32B & 113 & 1175 & 788 & 7813 & 728 & 7922 & 787 & 6511 & 786 & 9276 \\
Qwen 3 32B & 111 & 985 & 791 & 6099 & 730 & 6297 & 779 & 3647 & 786 & 4898 \\
R1 Qwen 32B & 113 & 600 & 790 & 5530 & 693 & 5310 & 709 & 2674 & 720 & 3118 \\
\bottomrule
\end{tabular}
\end{table}

\begin{table}[htbp]
    \centering
    \scriptsize
    \caption{Backtracking Next Token Frequency on sample rollouts from each dataset, split by mid-trace proposals and final proposals. We define the following words as backtracking words: \texttt{"Wait", ''But", "However", "Hmm", "Alternatively", "Maybe", "Another"}
    }
    \label{tab:backtrack-freq}
    \begin{tabular}{llrrrrrr}
    \toprule
    Model & Marker type & AIME 2026 & GSM8K & OMEGA-500 & GPQA-D & MMLU-Pro & All \\
    \midrule
    \multirow{2}{*}{Olmo 3 32B} & Final & 3.5\% & 0.5\% & 5.2\% & 0.3\% & 0.4\% & 1.6\% \\
     & Mid-trace & 76.9\% & 40.6\% & 68.6\% & 65.1\% & 65.2\% & 60.6\% \\
    \midrule
    \multirow{2}{*}{Qwen 3 32B} & Final & 3.6\% & 0.6\% & 5.6\% & 0.3\% & 0.5\% & 0.9\% \\
     & Mid-trace & 66.8\% & 37.8\% & 63.0\% & 64.2\% & 55.6\% & 52.2\% \\
    \midrule
    \multirow{2}{*}{Distill 32B} & Final & 4.4\% & 0.3\% & 2.5\% & 2.3\% & 2.2\% & 1.9\% \\
     & Mid-trace & 77.8\% & 39.6\% & 71.6\% & 78.8\% & 69.9\% & 61.8\% \\
    \bottomrule
    \end{tabular}
\end{table}

\begin{table}[htbp]
    \centering
    \caption{Test Marker Splits. \# M denotes number of markers. \# denotes corresponding number of unique rollouts that contained those markers}
    \label{tab:test_marker_splits}
    \scriptsize
\begin{tabular}{llrrrrrrrrrr}
\toprule
 &  & \multicolumn{2}{c}{AIME 2026} & \multicolumn{2}{c}{GSM8K} & \multicolumn{2}{c}{OMEGA-500} & \multicolumn{2}{c}{GPQA-D} & \multicolumn{2}{c}{MMLU-Pro} \\
\cmidrule(lr){3-4} \cmidrule(lr){5-6} \cmidrule(lr){7-8} \cmidrule(lr){9-10} \cmidrule(lr){11-12}
Model & Marker type & \# R & \# M & \# R & \# M & \# R & \# M & \# R & \# M & \# R & \# M \\
\midrule
\multirow{3}{*}{Olmo 3 32B} & Final & 41 & 41 & 50 & 50 & 50 & 50 & 50 & 50 & 50 & 50 \\
 & Mid-trace & 28 & 59 & 48 & 50 & 45 & 50 & 43 & 50 & 45 & 50 \\
 & All & 41 & 100 & 92 & 100 & 88 & 100 & 85 & 100 & 90 & 100 \\
\midrule
\multirow{3}{*}{Qwen 3 32B} & Final & 42 & 42 & 50 & 50 & 50 & 50 & 50 & 50 & 50 & 50 \\
 & Mid-trace & 27 & 58 & 44 & 50 & 30 & 50 & 40 & 50 & 45 & 50 \\
 & All & 42 & 100 & 90 & 100 & 59 & 100 & 84 & 100 & 89 & 100 \\
\midrule
\multirow{3}{*}{R1 Qwen 32B} & Final & 41 & 41 & 50 & 50 & 50 & 50 & 50 & 50 & 50 & 50 \\
 & Mid-trace & 28 & 59 & 41 & 50 & 40 & 50 & 43 & 50 & 40 & 50 \\
 & All & 41 & 100 & 86 & 100 & 86 & 100 & 83 & 100 & 85 & 100 \\
\bottomrule
\end{tabular}
\end{table}

\begin{table}[htbp]
  \centering
  \scriptsize
  \caption{Unique answers vs.\ answer proposals per rollout per model and dataset. \# unique is the number of distinct answers the judge confirmed in a rollout's reasoning trace; proposals is the mean $\pm$ std of all confirmed answer proposals, repeats included, over the $n$ rollouts in that row.}
  \label{tab:unique-vs-proposed}
  \begin{tabular}{lllrc}
    \toprule
    Model & Dataset & \# unique & $n$ & proposals \\
    \midrule
    \multirow{15}{*}{OLMo-3-32B} & \multirow{3}{*}{AIME 2026} & 1 & 107 & $9.7 \pm 5.5$ \\
     &  & 2 & 5 & $23.6 \pm 8.3$ \\
     &  & 3+ & 1 & $21.0 \pm 0.0$ \\
    \cmidrule(l){2-5}
     & \multirow{3}{*}{GSM8K} & 1 & 700 & $8.6 \pm 4.6$ \\
     &  & 2 & 70 & $18.6 \pm 14.8$ \\
     &  & 3+ & 18 & $26.7 \pm 16.2$ \\
    \cmidrule(l){2-5}
     & \multirow{3}{*}{OMEGA-500} & 1 & 643 & $9.9 \pm 6.7$ \\
     &  & 2 & 55 & $17.4 \pm 13.2$ \\
     &  & 3+ & 30 & $19.1 \pm 16.6$ \\
    \cmidrule(l){2-5}
     & \multirow{3}{*}{GPQA-D} & 1 & 620 & $6.7 \pm 5.1$ \\
     &  & 2 & 136 & $13.2 \pm 9.5$ \\
     &  & 3+ & 31 & $19.1 \pm 13.6$ \\
    \cmidrule(l){2-5}
     & \multirow{3}{*}{MMLU-Pro} & 1 & 633 & $8.8 \pm 8.6$ \\
     &  & 2 & 110 & $19.3 \pm 11.9$ \\
     &  & 3+ & 43 & $37.4 \pm 26.6$ \\
    \midrule
    \multirow{15}{*}{Qwen3-32B} & \multirow{3}{*}{AIME 2026} & 1 & 107 & $8.7 \pm 5.4$ \\
     &  & 2 & 3 & $13.3 \pm 6.8$ \\
     &  & 3+ & 1 & $13.0 \pm 0.0$ \\
    \cmidrule(l){2-5}
     & \multirow{3}{*}{GSM8K} & 1 & 712 & $7.0 \pm 3.9$ \\
     &  & 2 & 67 & $14.5 \pm 8.3$ \\
     &  & 3+ & 12 & $14.8 \pm 5.2$ \\
    \cmidrule(l){2-5}
     & \multirow{3}{*}{OMEGA-500} & 1 & 644 & $8.1 \pm 6.9$ \\
     &  & 2 & 59 & $12.4 \pm 8.9$ \\
     &  & 3+ & 27 & $13.5 \pm 7.4$ \\
    \cmidrule(l){2-5}
     & \multirow{3}{*}{GPQA-D} & 1 & 710 & $4.1 \pm 3.1$ \\
     &  & 2 & 57 & $8.4 \pm 6.0$ \\
     &  & 3+ & 12 & $19.2 \pm 10.5$ \\
    \cmidrule(l){2-5}
     & \multirow{3}{*}{MMLU-Pro} & 1 & 681 & $4.3 \pm 3.4$ \\
     &  & 2 & 70 & $9.7 \pm 5.6$ \\
     &  & 3+ & 35 & $36.1 \pm 17.5$ \\
    \midrule
    \multirow{15}{*}{R1-Qwen-32B} & \multirow{3}{*}{AIME 2026} & 1 & 97 & $4.7 \pm 3.7$ \\
     &  & 2 & 13 & $7.8 \pm 4.5$ \\
     &  & 3+ & 3 & $15.7 \pm 6.6$ \\
    \cmidrule(l){2-5}
     & \multirow{3}{*}{GSM8K} & 1 & 716 & $6.2 \pm 4.1$ \\
     &  & 2 & 58 & $14.8 \pm 13.1$ \\
     &  & 3+ & 16 & $12.6 \pm 10.0$ \\
    \cmidrule(l){2-5}
     & \multirow{3}{*}{OMEGA-500} & 1 & 577 & $6.2 \pm 5.8$ \\
     &  & 2 & 78 & $14.4 \pm 13.9$ \\
     &  & 3+ & 38 & $16.0 \pm 11.7$ \\
    \cmidrule(l){2-5}
     & \multirow{3}{*}{GPQA-D} & 1 & 627 & $3.2 \pm 2.5$ \\
     &  & 2 & 69 & $7.6 \pm 5.4$ \\
     &  & 3+ & 13 & $11.4 \pm 7.7$ \\
    \cmidrule(l){2-5}
     & \multirow{3}{*}{MMLU-Pro} & 1 & 634 & $3.2 \pm 2.9$ \\
     &  & 2 & 62 & $8.7 \pm 4.9$ \\
     &  & 3+ & 24 & $23.5 \pm 17.7$ \\
    \bottomrule
  \end{tabular}
\end{table}

\begin{table}[htbp]
\centering
\scriptsize
\setlength{\tabcolsep}{4pt}
\caption{Spearman $\rho$ between verbalized confidence (expected value of the 0 to 9 at each proposed-answer marker) and commitment (1 if the marker is the rollout's final answer, 0 if the model keeps reasoning). \emph{Pooled}: every marker as a point; rows spanning several datasets rank confidence within each dataset first. \emph{Within-rollout}: confidence centered on its rollout's mean, single-marker rollouts dropped. $^{*}$: the 95\% rollout-cluster bootstrap CI excludes 0. \revise{In parentheses is the max $\rho$ attainable against the binary commit label at that dataset subset's ratio of final to mid-trace markers.}}
\label{tab:vc_commit_correlations}
\resizebox{\linewidth}{!}{%
\begin{tabular}{ll rr c rrr c rrr}
\toprule
 & & & & & \multicolumn{3}{c}{Pooled $\rho$} & & \multicolumn{3}{c}{Within-rollout $\rho$} \\
\cmidrule{6-8} \cmidrule{10-12}
Model & Dataset & Rollouts & Markers & & All & Correct & Incorrect & & All & Correct & Incorrect \\
\midrule
\multirow{6}{*}{OLMo-3-32B-Think} & AIME 2026 & 113 & 1175 & & {$+0.21^{*}$\,{\scriptsize(.51)}} & {$+0.25^{*}$\,{\scriptsize(.50)}} & {$+0.07\phantom{^{*}}$\,{\scriptsize(.59)}} & & {$+0.30^{*}$\,{\scriptsize(.51)}} & {$+0.31^{*}$\,{\scriptsize(.50)}} & {$+0.23^{*}$\,{\scriptsize(.57)}} \\
 & OMEGA & 728 & 7922 & & {$+0.11^{*}$\,{\scriptsize(.50)}} & {$+0.16^{*}$\,{\scriptsize(.49)}} & {$+0.01\phantom{^{*}}$\,{\scriptsize(.52)}} & & {$+0.22^{*}$\,{\scriptsize(.50)}} & {$+0.26^{*}$\,{\scriptsize(.49)}} & {$+0.14^{*}$\,{\scriptsize(.52)}} \\
 & GSM8K & 788 & 7813 & & {$+0.28^{*}$\,{\scriptsize(.52)}} & {$+0.28^{*}$\,{\scriptsize(.54)}} & {$+0.20^{*}$\,{\scriptsize(.36)}} & & {$+0.25^{*}$\,{\scriptsize(.52)}} & {$+0.26^{*}$\,{\scriptsize(.54)}} & {$+0.22^{*}$\,{\scriptsize(.36)}} \\
 & GPQA-D & 787 & 6511 & & {$+0.02^{*}$\,{\scriptsize(.56)}} & {$+0.06^{*}$\,{\scriptsize(.56)}} & {$-0.02\phantom{^{*}}$\,{\scriptsize(.58)}} & & {$+0.16^{*}$\,{\scriptsize(.55)}} & {$+0.19^{*}$\,{\scriptsize(.55)}} & {$+0.09^{*}$\,{\scriptsize(.56)}} \\
 & MMLU-Pro & 786 & 9276 & & {$+0.10^{*}$\,{\scriptsize(.48)}} & {$+0.15^{*}$\,{\scriptsize(.51)}} & {$-0.03\phantom{^{*}}$\,{\scriptsize(.43)}} & & {$+0.14^{*}$\,{\scriptsize(.48)}} & {$+0.18^{*}$\,{\scriptsize(.50)}} & {$+0.04^{*}$\,{\scriptsize(.43)}} \\
\cmidrule{2-12}
 & \textit{All} & 3202 & 32697 & & {\boldmath$+0.13^{*}$\,{\scriptsize(.51)}} & {\boldmath$+0.18^{*}$\,{\scriptsize(.52)}} & {\boldmath$+0.00\phantom{^{*}}$\,{\scriptsize(.50)}} & & {\boldmath$+0.20^{*}$\,{\scriptsize(.51)}} & {\boldmath$+0.23^{*}$\,{\scriptsize(.52)}} & {\boldmath$+0.10^{*}$\,{\scriptsize(.49)}} \\
\midrule
\multirow{6}{*}{Qwen3-32B} & AIME 2026 & 111 & 985 & & {$+0.22^{*}$\,{\scriptsize(.55)}} & {$+0.32^{*}$\,{\scriptsize(.53)}} & {$+0.02\phantom{^{*}}$\,{\scriptsize(.65)}} & & {$+0.32^{*}$\,{\scriptsize(.54)}} & {$+0.33^{*}$\,{\scriptsize(.53)}} & {$+0.28^{*}$\,{\scriptsize(.61)}} \\
 & OMEGA & 730 & 6297 & & {$+0.14^{*}$\,{\scriptsize(.54)}} & {$+0.25^{*}$\,{\scriptsize(.54)}} & {$-0.00\phantom{^{*}}$\,{\scriptsize(.55)}} & & {$+0.27^{*}$\,{\scriptsize(.53)}} & {$+0.29^{*}$\,{\scriptsize(.53)}} & {$+0.24^{*}$\,{\scriptsize(.54)}} \\
 & GSM8K & 791 & 6099 & & {$+0.31^{*}$\,{\scriptsize(.58)}} & {$+0.31^{*}$\,{\scriptsize(.59)}} & {$+0.27^{*}$\,{\scriptsize(.47)}} & & {$+0.26^{*}$\,{\scriptsize(.58)}} & {$+0.26^{*}$\,{\scriptsize(.59)}} & {$+0.26^{*}$\,{\scriptsize(.47)}} \\
 & GPQA-D & 779 & 3647 & & {$+0.04^{*}$\,{\scriptsize(.71)}} & {$+0.11^{*}$\,{\scriptsize(.69)}} & {$-0.02\phantom{^{*}}$\,{\scriptsize(.75)}} & & {$+0.35^{*}$\,{\scriptsize(.68)}} & {$+0.36^{*}$\,{\scriptsize(.67)}} & {$+0.31^{*}$\,{\scriptsize(.70)}} \\
 & MMLU-Pro & 786 & 4898 & & {$+0.16^{*}$\,{\scriptsize(.64)}} & {$+0.17^{*}$\,{\scriptsize(.67)}} & {$+0.03\phantom{^{*}}$\,{\scriptsize(.54)}} & & {$+0.22^{*}$\,{\scriptsize(.61)}} & {$+0.23^{*}$\,{\scriptsize(.65)}} & {$+0.21^{*}$\,{\scriptsize(.52)}} \\
\cmidrule{2-12}
 & \textit{All} & 2647 & 17255 & & {\boldmath$+0.18^{*}$\,{\scriptsize(.62)}} & {\boldmath$+0.22^{*}$\,{\scriptsize(.63)}} & {\boldmath$+0.03\phantom{^{*}}$\,{\scriptsize(.61)}} & & {\boldmath$+0.27^{*}$\,{\scriptsize(.61)}} & {\boldmath$+0.27^{*}$\,{\scriptsize(.61)}} & {\boldmath$+0.25^{*}$\,{\scriptsize(.58)}} \\
\midrule
\multirow{6}{*}{R1-Distill-Qwen-32B} & AIME 2026 & 113 & 600 & & {$+0.17^{*}$\,{\scriptsize(.68)}} & {$+0.29^{*}$\,{\scriptsize(.68)}} & {$+0.01\phantom{^{*}}$\,{\scriptsize(.67)}} & & {$+0.23^{*}$\,{\scriptsize(.65)}} & {$+0.40^{*}$\,{\scriptsize(.66)}} & {$-0.05\phantom{^{*}}$\,{\scriptsize(.64)}} \\
 & OMEGA & 693 & 5310 & & {$+0.12^{*}$\,{\scriptsize(.58)}} & {$+0.24^{*}$\,{\scriptsize(.60)}} & {$+0.01\phantom{^{*}}$\,{\scriptsize(.57)}} & & {$+0.22^{*}$\,{\scriptsize(.57)}} & {$+0.28^{*}$\,{\scriptsize(.59)}} & {$+0.15^{*}$\,{\scriptsize(.55)}} \\
 & GSM8K & 790 & 5530 & & {$+0.35^{*}$\,{\scriptsize(.61)}} & {$+0.35^{*}$\,{\scriptsize(.61)}} & {$+0.33^{*}$\,{\scriptsize(.49)}} & & {$+0.39^{*}$\,{\scriptsize(.61)}} & {$+0.41^{*}$\,{\scriptsize(.61)}} & {$+0.22^{*}$\,{\scriptsize(.49)}} \\
 & GPQA-D & 709 & 2674 & & {$+0.01\phantom{^{*}}$\,{\scriptsize(.76)}} & {$+0.06^{*}$\,{\scriptsize(.76)}} & {$-0.04\phantom{^{*}}$\,{\scriptsize(.78)}} & & {$+0.19^{*}$\,{\scriptsize(.72)}} & {$+0.29^{*}$\,{\scriptsize(.72)}} & {$+0.01\phantom{^{*}}$\,{\scriptsize(.73)}} \\
 & MMLU-Pro & 720 & 3118 & & {$+0.14^{*}$\,{\scriptsize(.73)}} & {$+0.17^{*}$\,{\scriptsize(.78)}} & {$-0.09^{*}$\,{\scriptsize(.61)}} & & {$+0.19^{*}$\,{\scriptsize(.67)}} & {$+0.33^{*}$\,{\scriptsize(.73)}} & {$-0.04\phantom{^{*}}$\,{\scriptsize(.57)}} \\
\cmidrule{2-12}
 & \textit{All} & 3025 & 17232 & & {\boldmath$+0.17^{*}$\,{\scriptsize(.66)}} & {\boldmath$+0.23^{*}$\,{\scriptsize(.67)}} & {\boldmath$+0.00\phantom{^{*}}$\,{\scriptsize(.62)}} & & {\boldmath$+0.25^{*}$\,{\scriptsize(.63)}} & {\boldmath$+0.33^{*}$\,{\scriptsize(.65)}} & {\boldmath$+0.08^{*}$\,{\scriptsize(.59)}} \\
\midrule
\multicolumn{2}{l}{\textit{All models}} & 8874 & 67184 & & {\boldmath$+0.15^{*}$\,{\scriptsize(.59)}} & {\boldmath$+0.21^{*}$\,{\scriptsize(.59)}} & {\boldmath$+0.01\phantom{^{*}}$\,{\scriptsize(.57)}} & & {\boldmath$+0.22^{*}$\,{\scriptsize(.57)}} & {\boldmath$+0.26^{*}$\,{\scriptsize(.58)}} & {\boldmath$+0.13^{*}$\,{\scriptsize(.54)}} \\
\bottomrule
\end{tabular}}
\end{table}

\FloatBarrier

\section{Vector Consistency}
\label{app:confidence_consistency}

\paragraph{Verbalized Confidence Prompts}
We test the robustness of the confidence extraction method to various verbalized confidence prompts on AIME 2026.
We use the following prompts:
\begin{itemize}
    \item Range 9: My confidence in this answer, on a scale from 0 to 9, is 
    \item Range 100: My confidence in this answer, on a scale from 1 to 100, is 
    \item Range 3: My confidence in this answer, chosen from ``low", ``medium", ``high", is '' 
\end{itemize}

We compare the spearman rank correlation between the rankings of the markers from low to high verbalized confidence per pair of prompts. Across all three models, the rank correlation averaged across the three pairs of prompts is at least 0.932. This demonstrates that each answer proposal marker ranks similarly in its verbalized confidence across the three prompts.
Similarly, we take the centered kernel alignment \cite{kornblith2019similarityneuralnetworkrepresentations} similarity between the representations of the final token extracted between each pair of prompts.
While CKA is typically used to measure the similarities of representations between two models while keeping the input the same, an equivalent setup is to measure the similarities of representations between two different prompts while keeping the model and the prefix tokens (prompt and model response) the same.
Across all models, the CKA averaged across the three pars of prompts is at least 0.876. See \autoref{tab:range-agreement}.

\begin{figure}
    \centering
    \begin{subfigure}[b]{0.9\linewidth}
        \centering
        \includegraphics[width=\linewidth]{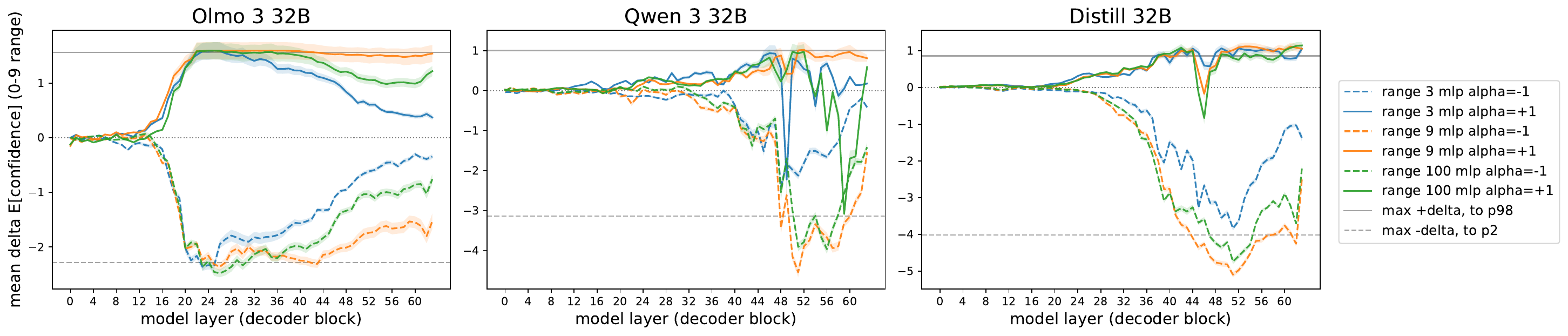}
        \caption{Verbalized confidence steering on AIME 2026 across verbalized confidence prompts}
        \label{fig:vc_steer_by_vc_prompt}
    \end{subfigure}
    \begin{subfigure}[b]{0.9\linewidth}
        \centering
        \includegraphics[width=\linewidth]{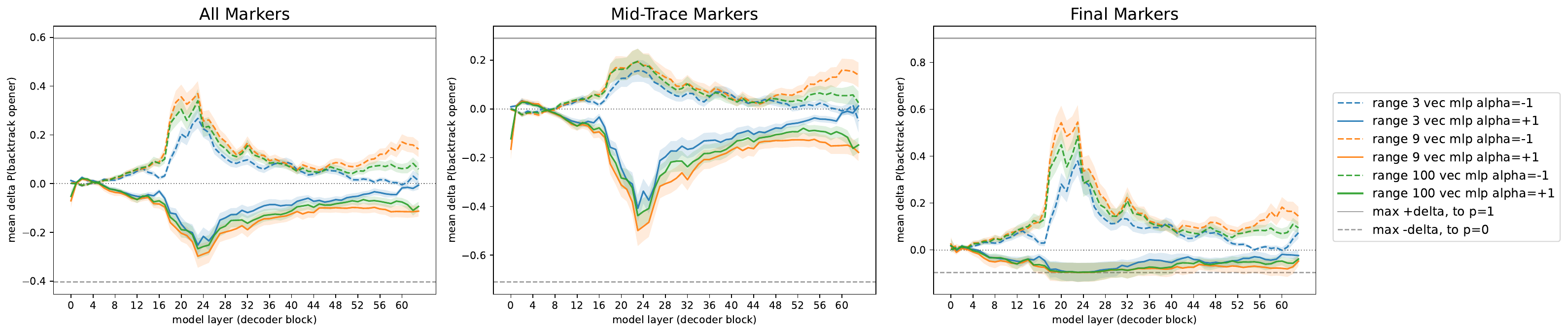}
        \caption{Olmo 3 32B $p_\text{bt}$ steering on AIME 2026 across verbalized confidence prompts}
        \label{fig:commit_steer_by_vc_prompt}
    \end{subfigure}
    \caption{Steering Verbalized Confidence and $p_\text{bt}$ for Range 3, Range 9, and Range 100 confidence directions.
    For visual clarity, we only show results for steering at post-MLP residual stream sites.}
    \label{fig:vc_prompts}
\end{figure}

Lastly, we measure how well each confidence direction can steer verbalized confidence with a range of 9 on AIME 2026.
\autoref{fig:vc_steer_by_vc_prompt} shows the steering performance of the DIM vectors across all layers. 
The Difference in Means direction obtained directly from the range 9 scale saturates the positive headroom and remains close to the negative headroom from the middle layers onward. The other two DIM directions perform equally well in the middle to late layers, with performance tapering near the end, likely due to the lexical differences in the prompts finally taking over.
We also get the performance of the range 100 and range 3 vectors on AIME 2026, and find similar performance to the range 9 vector, so we stick with range 9 for the rest of the experiments, as shown in \autoref{fig:commit_steer_by_vc_prompt}.

\begin{table}[htbp]
    \centering
    \begin{tabular}{lcccccc}
        \toprule
        & \multicolumn{2}{c}{Olmo 3 32B} & \multicolumn{2}{c}{Qwen 3 32B} & \multicolumn{2}{c}{Distill 32B} \\
        \cmidrule(lr){2-3} \cmidrule(lr){4-5} \cmidrule(lr){6-7}
        & Spearman & CKA & Spearman & CKA & Spearman & CKA \\
        \midrule
        Range 9, Range 100 & 0.978 & 0.967 & 0.969 & 0.961 & 0.979 & 0.958 \\
        Range 3, Range 100 & 0.910 & 0.835 & 0.962 & 0.856 & 0.942 & 0.962 \\
        Range 3, Range 9   & 0.907 & 0.827 & 0.958 & 0.852 & 0.939 & 0.969 \\
        \midrule
        Average            & 0.932 & 0.876 & 0.963 & 0.890 & 0.953 & 0.963 \\
        \bottomrule
    \end{tabular}
    \caption{Spearman correlation and CKA between representations across digit ranges.}
    \label{tab:range-agreement}
\end{table}

\FloatBarrier

\section{Ablation Experiments}
\label{app:ablation_misc}

\subsection{Probability DiM}
\label{app:ablation_prob}
$v_\text{conf}$ being able to steer $p_\text{bt}$ may trivially be explained as the verbalized confidence prompt simply decoding $p_\text{bt}$ from the last token position of the answer proposal.
To separate this effect from $v_\text{conf}$, we construct $v_\text{prob}$ --- a DiM direction separating high and low $p_\text{bt}$ classes --- and measure its performance on VC steering.
Since a histogram of the proposal markers binned by $p_\text{bt}$ is bimodal, concentrated near 0-0.1 and 0.9-1, we separate markers into high and low probability classes based on a threshold of 0.5.
Each model's $v_\text{prob}$ is computed from the same dataset that produced the best $v_\text{conf}$.
As shown in \autoref{tab:st_steer_best}, the results of this probability direction on steering $p_\text{bt}$ are the best among the three directions.
\revise{This is expected because $v_\text{prob}$ is constructed specifically from the distance between markers with high and low $p_\text{bt}$.}
However, it performs poorly on steering verbalized confidence, slightly better than $v_\text{commit}$ for Distill 32B and slightly worse for Olmo 3 32B and Qwen 3 32B.
This demonstrates that the latent confidence representation is not just a trivial decoder for answer proposal next token probability.

\subsection{Position Ablation}
\label{app:ablation_position}
In \cref{sec:3dimconstruction}, we established a positive correlation between verbalized confidence at a proposal and whether the proposal is a mid-trace or final marker.
Intuitively, within a single rollout, the model's verbalized confidence increases as the model grows in confidence as it explores the possible solution space and accumulates evidence.
Given this correlation, selecting high and low confidence markers to construct the direction may amount to noisily selecting late-position vs.\ early-position markers, in which case $v_\text{conf}$ would be representing position rather than confidence.

To ablate positional information, we construct a confidence direction, $v_\text{conf-pos}$, such that the markers selected for the high and low confidence classes share the same positional distribution.
We index the markers of rollout $r$ by their order in the text, $j \in \{0,\dots,n_r-1\}$, and define the normalized position $\pi = j/(n_r-1)$, with $\pi \equiv 1$ for single-marker rollouts.
We define the \emph{positional drift} of a fitted direction as the difference in the mean normalized position between its two classes, $\bar\pi_\text{high} - \bar\pi_\text{low}$, measured over the markers that entered the fit. $0$ indicates the classes are drawn from the same place in the trace.
We then assign the markers in the train split into four positional bins: first marker, early half, late half, and final marker. If a rollout only contains one marker, it is assigned as a final marker. Thus, each marker is assigned a positional bin and a binary confidence label.

Let $H_i$ and $L_i$ be the number of high and low confidence markers in position bin $i$. 
When selecting markers, bin $i$ can supply at most $U_i=\min(H_i,L_i)$ markers of each class since supplying more than $U_i$ would mean one class contains more markers from that bin than the other. From bin $i$, we draw $k_i = \text{round}(N p_i)$ samples of each class with $p_i = U_i / \sum_j U_j$ and $N = \min(100, \sum_i U_i)$. Intuitively, bins that contain fewer markers should also provide fewer markers to construct the DiM vector. The $k_i$ are calculated based off the train split and fixed before sampling.
For each model, we construct $v_\text{conf-pos}$ from the dataset split of the best performing $v_\text{conf}$. \autoref{tab:posmatched-gap} shows the positional drift of $v_\text{conf-pos}$ and $v_\text{conf}$.

Enforcing a shared positional distribution based on discrete bins eliminates positional drift between bins by construction. However, slight positional bias within the early half and late half bins persists since the classes may still differ in position inside a given bin.
Empirically this residual is small: $v_\text{conf-pos}$ on average drops 83\% of the positional drift relative to $v_\text{conf}$ (\autoref{tab:posmatched-gap}), while retaining 89.2\% of $v_\text{conf}$'s steering gap on $p_\text{bt}$ (\autoref{tab:st_steer_best}).
The confidence direction therefore cannot be explained solely as a position detector.

\begin{table}[htbp]
\centering
\caption{Position drift of $v_\text{conf}$ and $v_\text{conf-pos}$. $0$ means the two classes are drawn from the same relative proposal positions in the trace.}
\label{tab:posmatched-gap}
\begin{tabular}{llrrr}
\toprule
Model & Dataset & $v_\text{conf}$ & $v_\text{conf-pos}$ & Reduction \\
\midrule
Olmo-3-32B-Think & AIME-2026 & $+0.267$ & $+0.049$ & 82\% \\
Qwen3-32B & AIME-2026 & $+0.269$ & $+0.068$ & 75\% \\
R1-Distill-Qwen-32B & GSM8K & $+0.264$ & $+0.020$ & 92\% \\
\midrule
\textit{mean} & & $+0.266$ & $+0.046$ & 83\% \\
\bottomrule
\end{tabular}
\end{table}

\FloatBarrier

\section{Single Token Steering Details}
\label{app:st_steer_details}

\paragraph{Steering Details}
For the validation sweep in \cref{sec:3dimconstruction}, we sample 25 final and 25 mid-trace markers for a total of 50 markers in validation set. We sweep with $\alpha \in \{-1, -0.5, -0.25, 0.25, 0.5, 1\}$. Since the residual stream norm of a model gradually increases across layers, we set $\alpha$ to be in units of the steering site's average residual norm, so the overall intervention is the product of the $\alpha$ and residual norm scalars with the unit direction $v_\text{conf}$.

\textbf{Steering $p_\text{bt}$}
We follow the same procedure as described in \cref{sec:3dimconstruction}.
Since the AIME 2026 has at most 120 final markers (30 prompts $\times$ 4 rollouts, including judge pipeline filtering) split across train, validation, and test, there are fewer than 100 final proposal markers in the train set. Here, we select all available markers for $M^+$, and keep $|M^-|=100$.

\begin{table}[htbp]
\centering
\scriptsize
\caption{$p_\text{bt}$ and VC steering with each model's best $v_\text{commit}$ and $v_\text{conf}$ on the marker test set. We also test $v_\text{prob}$ and $v_\text{conf-pos}$ from App. \cref{app:ablation_misc}, which are fit on the best dataset for $v_\text{conf}$. Gap is the absolute distance between the positive- and negative-steering effects. See \autoref{tab:steer-test} for full results on $v_\text{commit}$ and $v_\text{conf}$.}
\setlength{\tabcolsep}{4pt}
\begin{tabular}{llllrrrllrrr}
\toprule
 &  & \multicolumn{5}{c}{$\Delta p_\text{bt}$} & \multicolumn{5}{c}{$\Delta$ VC} \\
\cmidrule(lr){3-7} \cmidrule(lr){8-12}
Model & Vector & Dataset & Site & Negative & Positive & Gap & Dataset & Site & Negative & Positive & Gap \\
\midrule
\multirow{4}{*}{Olmo 3 32B} & $v_\text{commit}$ & GPQA-D & 43.attn & +0.258 & -0.298 & 0.556 & GSM8K & 24.attn & -0.471 & +0.609 & 1.080 \\
 & $v_\text{conf}$ & AIME 2026 & 23.attn & +0.361 & -0.224 & 0.585 & GPQA-D & 23.attn & -3.502 & +1.215 & 4.717 \\
 \cmidrule{2-12}
 & $v_\text{prob.}$ & AIME 2026 & 28.attn & +0.486 & -0.300 & 0.786 & GPQA-D & 23.attn & -0.111 & +0.294 & 0.405 \\
 & $v_\text{conf-pos.}$ & AIME 2026 & 23.attn & +0.305 & -0.201 & 0.506 & GPQA-D & 23.attn & -3.467 & +1.221 & 4.688 \\
\midrule
\multirow{4}{*}{Qwen 3 32B} & $v_\text{commit}$ & GPQA-D & 51.mlp & +0.399 & -0.242 & 0.641 & GSM8K & 48.attn & -0.627 & +0.595 & 1.222 \\
 & $v_\text{conf}$ & AIME 2026 & 47.attn & +0.423 & -0.222 & 0.645 & MMLU-Pro & 51.attn & -3.284 & +1.002 & 4.286 \\
 \cmidrule{2-12}
 & $v_\text{prob.}$ & AIME 2026 & 47.attn & +0.681 & -0.273 & 0.954 & MMLU-Pro & 52.attn & -0.245 & +0.481 & 0.726 \\
 & $v_\text{conf-pos.}$ & AIME 2026 & 47.attn & +0.396 & -0.165 & 0.561 & MMLU-Pro & 49.attn & -1.917 & +0.816 & 2.733 \\
\midrule
\multirow{4}{*}{Distill 32B} & $v_\text{commit}$ & MMLU-Pro & 46.attn & +0.432 & -0.382 & 0.814 & AIME 2026 & 48.mlp & -0.290 & +0.273 & 0.563 \\
 & $v_\text{conf}$ & GSM8K & 46.attn & +0.403 & -0.341 & 0.744 & AIME 2026 & 51.attn & -5.198 & +0.891 & 6.089 \\
 \cmidrule{2-12}
 & $v_\text{prob.}$ & GSM8K & 43.attn & +0.608 & -0.337 & 0.945 & AIME 2026 & 45.attn & -0.556 & +0.523 & 1.079 \\
 & $v_\text{conf-pos.}$ & GSM8K & 47.attn & +0.388 & -0.307 & 0.695 & AIME 2026 & 51.attn & -5.251 & +0.883 & 6.134 \\
\bottomrule
\end{tabular}
\label{tab:st_steer_best}
\end{table}

\paragraph{Best steer site selection criteria}
Given a model with $L$ layers, we obtain a set of $2L$ directions $V = [v_1, v_2, \ldots, v_{2L}]$ from the DIM formulation, one direction from each post-attention and post-mlp residual across all layers. 
We ignore the first 15\% and last 15\% of layers, and select the best site $s$ from the middle 70\% of layers.
Let $E_{+}, E_{-}$ be the maximum effect of positive and negative steering from the candidate layers. Let $e^s_{+}, e^s_-$ be the effect of positive and negative steering of $v_s$. 
We select the best site using the criterion $C =\max_{s, \alpha}\left[\min(\frac{e^s_+}{E_+}, \tau) + \min(\frac{e^s_-}{E_-},\tau)\right]$ with threshold $\tau$. In other words, we select the steering layers and $\alpha$s that maximize the steering effect, clipped to $\tau$. If multiple $s,a$ satisfy the criterion, we select the earliest steering site and smallest $\alpha$ from all $s, \alpha$ combinations that satisfy the criteria $C$.

When steering $p_\text{bt}$ with the commitment direction or steering verbalized confidence with the confidence direction, steering performance rises sharply in the middle layers and stays nearly constant towards the final layers.
In these cases, to avoid selecting an arbitrarily late layer, we set threshold $\tau=0.9$.
When steering $p_\text{bt}$ with the confidence direction or steering verbalized confidence with the commitment direction, performance peaks in the middle layers and degrades towards the final layers, so we set a threshold $\tau=1$.

\begin{figure}[htbp]
    \centering
    \includegraphics[width=0.32\linewidth]{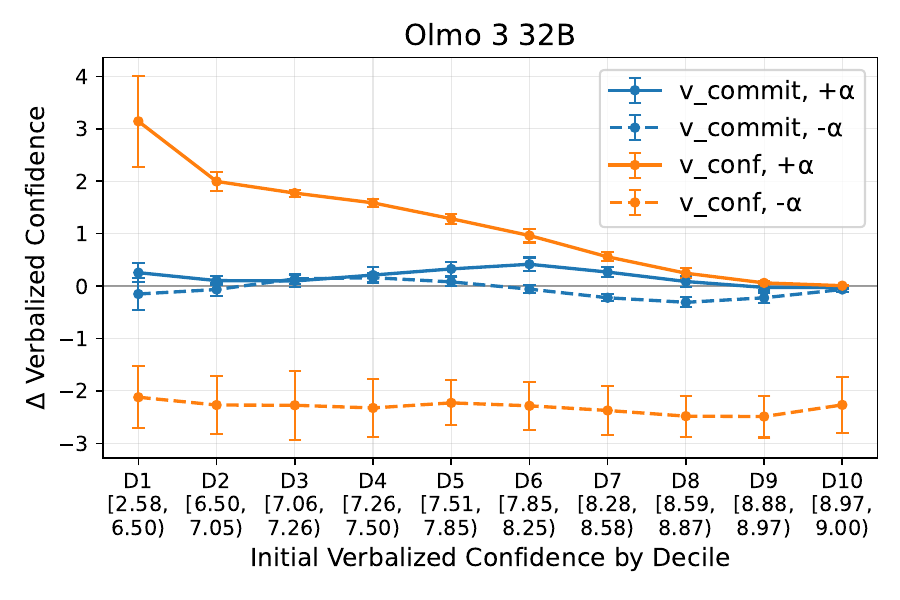}
    \includegraphics[width=0.32\linewidth]{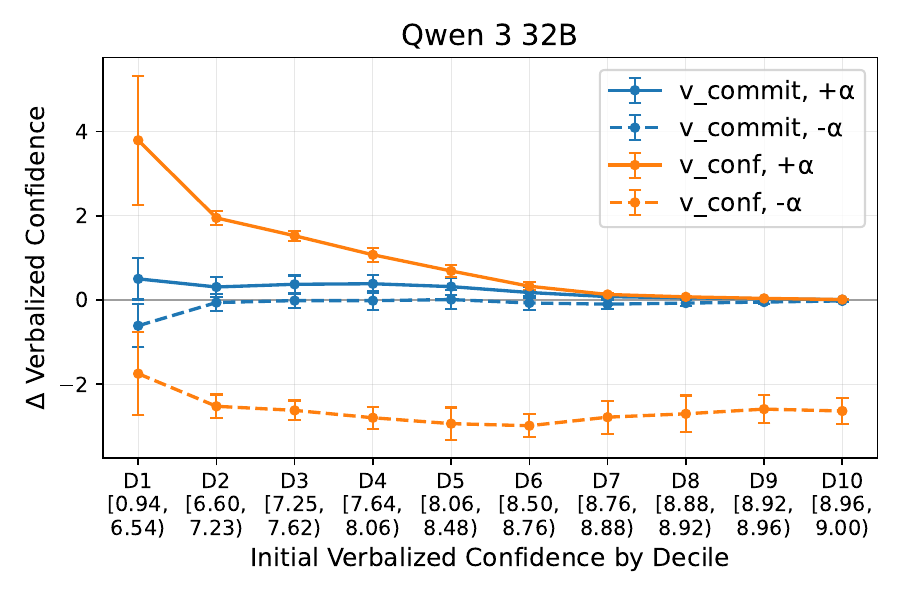}
    \includegraphics[width=0.32\linewidth]{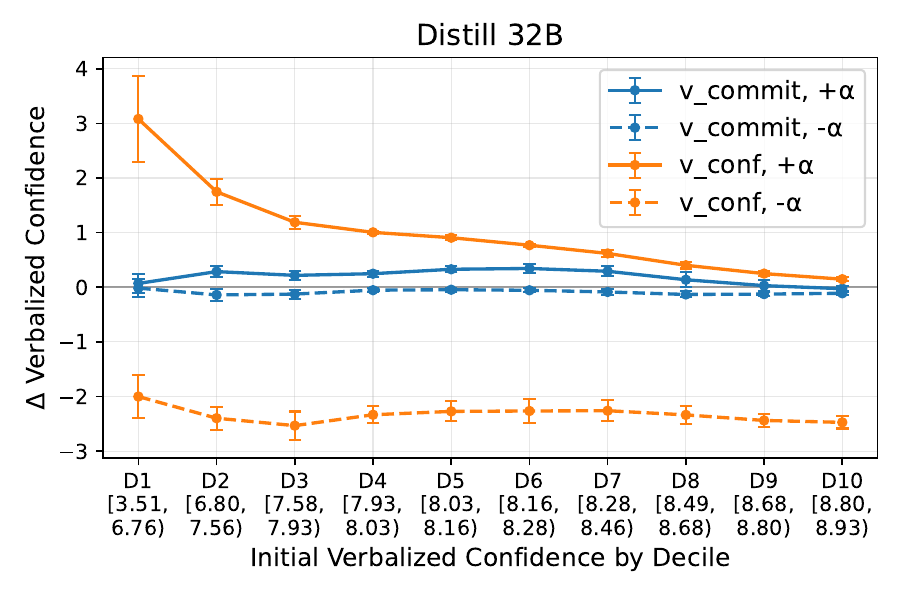}
    \caption{\revise{Steering results on the pooled test set, conditioned on the initial verbalized confidence of the answer proposal. Results are grouped by decile.}}
    \label{fig:steer_vc_deciles}
\end{figure}

\revise{
\paragraph{Verbalized Confidence Steering Analysis}
\label{app:st_steer_details:vc_steer_anal}
In \cref{sec:4st_steer}, we show that $v_\text{commit}$ cannot transfer to steering verbalized confidence. Astute readers may notice in \autoref{tab:st_steer_best_small} that there is an asymmetry between positive and negative steering. 
For negative steering, $v_\text{commit}$ obtains scores that are only 6-19\% of $v_\text{conf}$'s.
On the other hand, for positive steering, $v_\text{commit}$ obtains scores that are 31-59\% of $v_\text{conf}$'s, which is not insignificant. At face value, $v_\text{commit}$'s positive steering results seem to weaken our claim, but this can actually be explained as an artifact of the verbalized confidence range having a max value of 9.
In \autoref{fig:steer_vc_deciles}, we stratify the steering scores by separating the results into decile bins based on the answer proposal's initial confidence. D1 corresponds to the answer proposals with the bottom 10\% of initial verbalized confidences, and D10 corresponds to the top 10\%.
$v_\text{commit}$'s mean $|\Delta VC| < 1$ across all deciles for both positive and negative steering, while $v_\text{conf}$'s mean $|\Delta VC| > 2$ at almost all deciles for negative steering and the early deciles for positive steering. 
Since, the verbalized confidence scale is from 0 to 9, positive steering at higher initial confidences (D6 and above) hit a ceiling, which artificially deflates those scores. However, at lower initial confidences where ceiling for positive steering is higher (D5 and below), $v_\text{conf}$ strongly outperforms $v_\text{commit}$. Aggregating the steering results across deciles hides this detail, and makes $v_\text{commit}$'s positive steering scores seem closer than they actually are to $v_\text{conf}$'s scores.
}

\begin{figure}[htbp]
    \centering
    \includegraphics[width=0.95\linewidth]{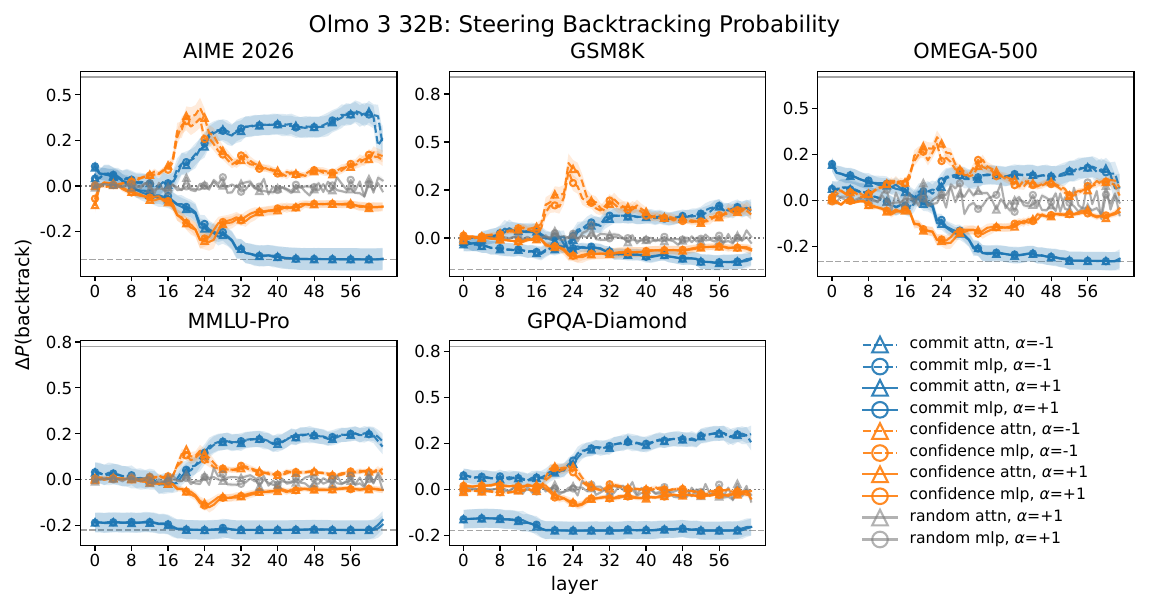}
    \includegraphics[width=0.95\linewidth]{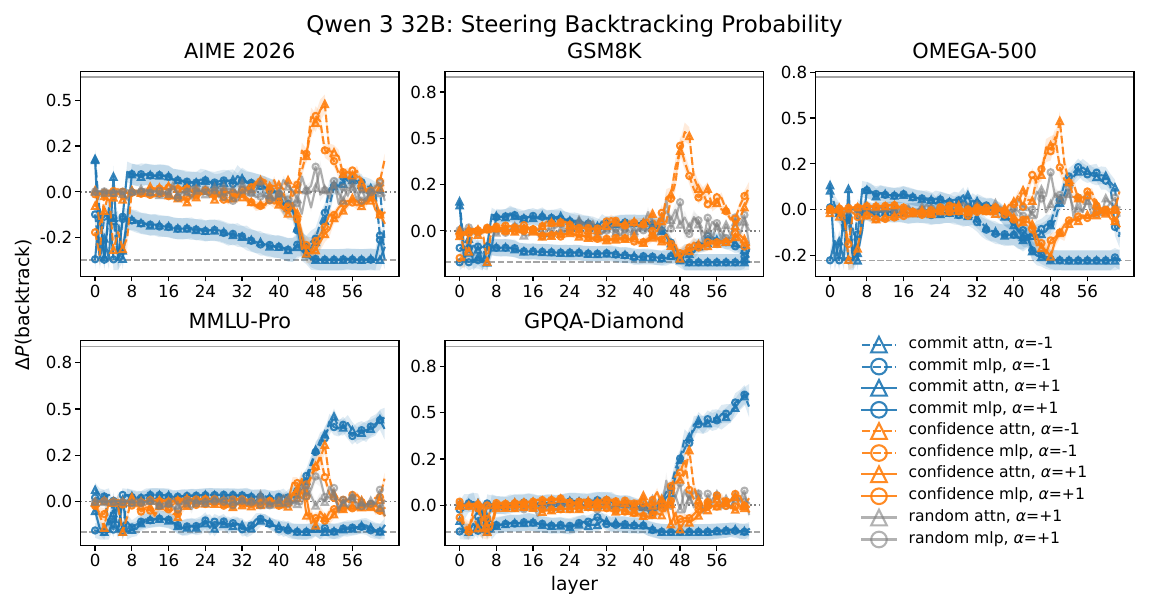}
    \includegraphics[width=0.95\linewidth]{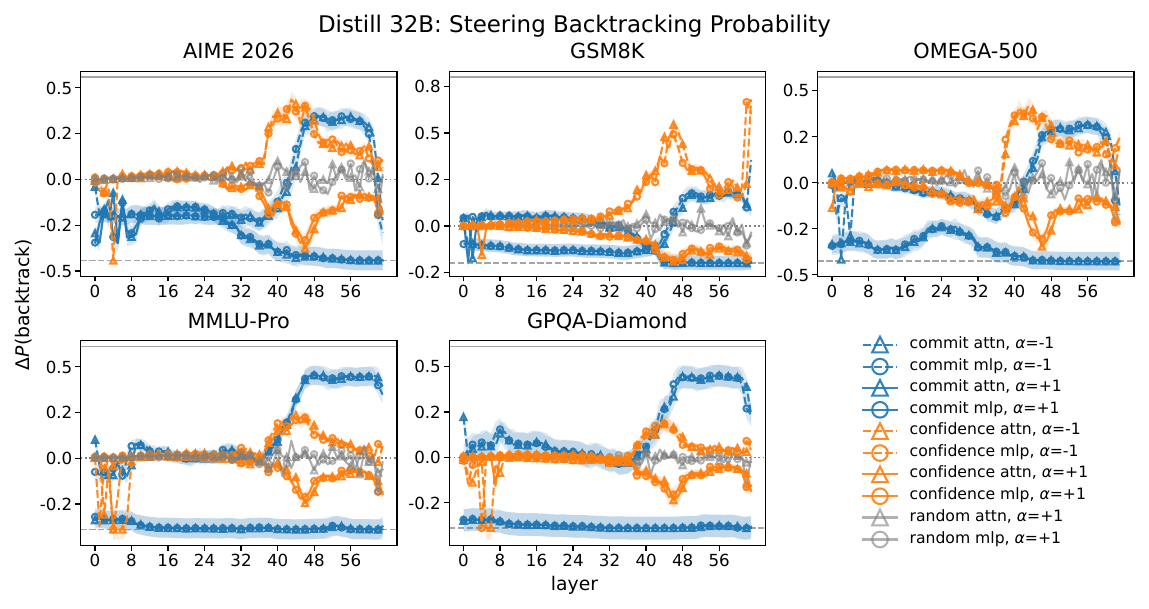}
    \caption{Steering Backtracking Token Results on Olmo 3 32B, Qwen 3 32B, and Distill 32B.}
    \label{fig:st_commit_steer_all}
\end{figure}

\begin{figure}[htbp]
    \centering
    \includegraphics[width=0.95\linewidth]{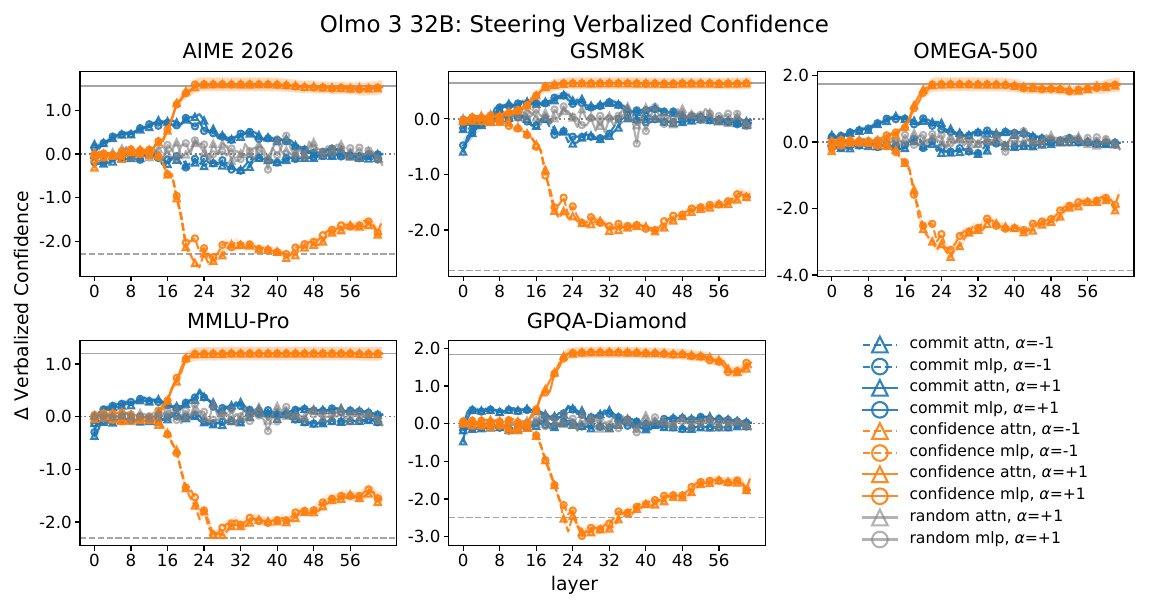}
    \includegraphics[width=0.95\linewidth]{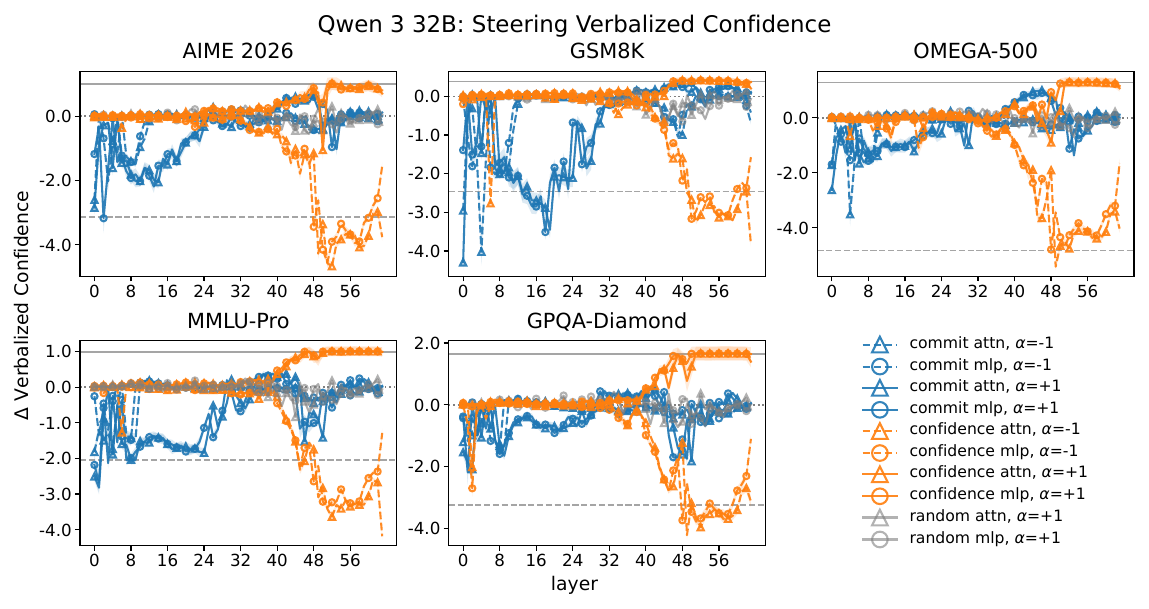}
    \includegraphics[width=0.95\linewidth]{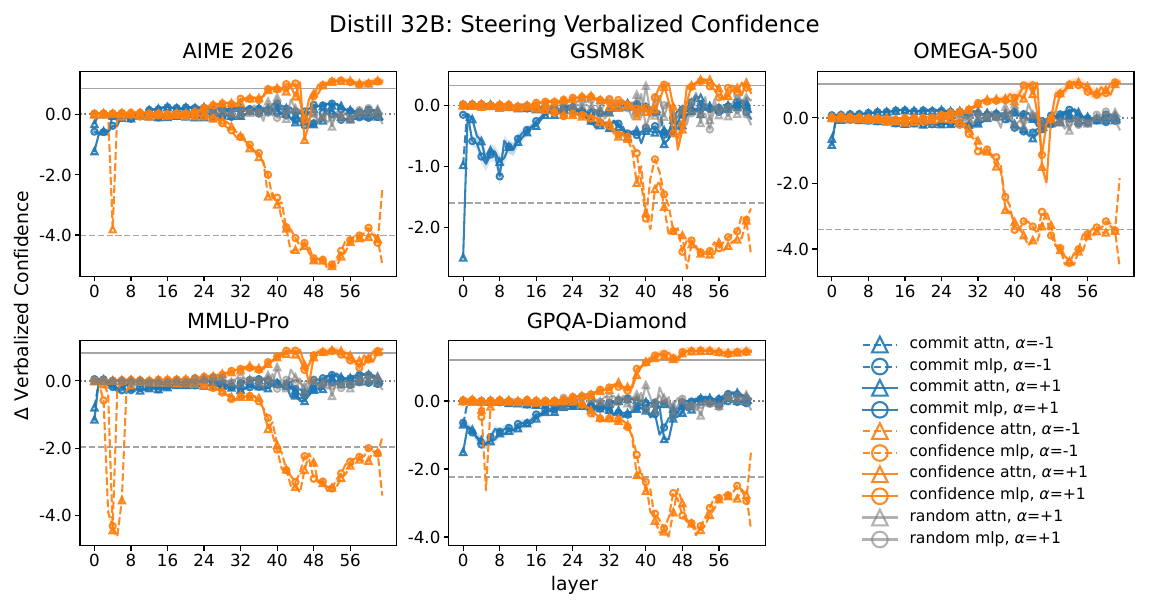}
    \caption{Confidence Steering Results on Olmo 3 32B, Qwen 3 32B, and Distill 32B.}
    \label{fig:st_conf_steer_all}
\end{figure}

\begin{table*}[t]
\centering
\small
\caption{Steering on the standard test set. $\Delta P(\mathrm{backtrack})$ is measured at backtrack/commit marker positions; $\Delta E$ is the change in expected confidence over the confidence probe positions, reported overall and per confidence tercile (the position axis is collapsed by an $n$-weighted average over the three position roles). Negative/Positive denote the sign of the steering coefficient $\alpha$; $|\alpha|=1$, except that $^{*}$ next to a site marks $|\alpha|=0.5$ and $^{\#}$ marks $|\alpha|=0.25$. Each task selects its own best site. Within each model and vector type, \textbf{bold} and \underline{underline} marks the first and second strongest score.
\label{tab:steer-test}
}
\setlength{\tabcolsep}{3pt}
\resizebox{\textwidth}{!}{%
\begin{tabular}{llllrrrrrrlrrrrrrrr}
\toprule
 &  &  & \multicolumn{7}{c}{$\Delta$P(backtrack)} & \multicolumn{9}{c}{$\Delta$E} \\
\cmidrule(lr){4-10} \cmidrule(lr){11-19}
 &  &  & Site & \multicolumn{3}{c}{Negative} & \multicolumn{3}{c}{Positive} & Site & \multicolumn{4}{c}{Negative} & \multicolumn{4}{c}{Positive} \\
\cmidrule(lr){5-7} \cmidrule(lr){8-10} \cmidrule(lr){12-15} \cmidrule(lr){16-19}
Model & Vector & Dataset &  & All & Backtrack & Commit & All & Backtrack & Commit &  & All & High & Medium & Low & All & High & Medium & Low \\
\midrule
\multirow[t]{10}{*}{Olmo 3 32B} & \multirow[t]{5}{*}{Commit} & AIME 2026 & 35.mlp & \underline{+0.209} & \textbf{+0.070} & +0.359 & \underline{-0.290} & \underline{-0.487} & \textbf{-0.078} & 20.mlp & \underline{-0.216} & -0.131 & -0.121 & \underline{-0.472} & \textbf{+0.623} & \underline{+0.277} & \textbf{+0.803} & \textbf{+0.949} \\
 &  & GSM8K & 53.mlp & +0.127 & -0.019 & +0.284 & -0.287 & -0.483 & \textbf{-0.078} & 24.attn & \textbf{-0.471} & \textbf{-0.452} & \textbf{-0.427} & \textbf{-0.560} & \underline{+0.609} & \textbf{+0.286} & \underline{+0.780} & \underline{+0.909} \\
 &  & OMEGA-500 & 39.attn & +0.049 & -0.089 & +0.197 & -0.265 & -0.439 & \textbf{-0.078} & 24.attn$^{*}$ & -0.186 & -0.162 & \underline{-0.166} & -0.251 & +0.269 & +0.174 & +0.331 & +0.343 \\
 &  & GPQA-D & 43.attn & \textbf{+0.258} & \underline{+0.056} & \textbf{+0.476} & \textbf{-0.298} & \textbf{-0.502} & \textbf{-0.078} & 26.mlp & -0.072 & \underline{-0.173} & +0.040 & -0.051 & +0.172 & +0.144 & +0.199 & +0.181 \\
 &  & MMLU-Pro & 43.attn & +0.204 & +0.018 & \underline{+0.405} & -0.286 & -0.480 & \textbf{-0.078} & 24.attn$^{*}$ & -0.064 & -0.078 & -0.010 & -0.111 & +0.211 & +0.134 & +0.242 & +0.296 \\
\cmidrule(lr){2-19}
 & \multirow[t]{5}{*}{Confidence-9} & AIME 2026 & 23.attn & \textbf{+0.361} & \textbf{+0.302} & \textbf{+0.425} & \textbf{-0.224} & \textbf{-0.365} & \textbf{-0.074} & 22.attn & -2.310 & -2.409 & -2.204 & -2.288 & +1.162 & +0.369 & +1.231 & +2.349 \\
 &  & GSM8K & 25.attn & \underline{+0.276} & \underline{+0.254} & \underline{+0.298} & \underline{-0.213} & \underline{-0.344} & \underline{-0.073} & 25.attn & -1.538 & -1.961 & -1.400 & -1.035 & \textbf{+1.230} & \textbf{+0.378} & \textbf{+1.276} & \textbf{+2.539} \\
 &  & OMEGA-500 & 23.attn & +0.242 & +0.229 & +0.255 & -0.169 & -0.264 & -0.066 & 23.attn & \underline{-3.280} & \underline{-3.786} & \underline{-3.088} & \textbf{-2.710} & \underline{+1.229} & \textbf{+0.378} & \underline{+1.275} & \underline{+2.536} \\
 &  & GPQA-D & 25.attn & +0.124 & +0.136 & +0.111 & -0.113 & -0.175 & -0.048 & 23.attn & \textbf{-3.502} & \textbf{-4.141} & \textbf{-3.356} & \underline{-2.663} & +1.215 & \underline{+0.377} & +1.268 & +2.494 \\
 &  & MMLU-Pro & 23.attn & +0.201 & +0.195 & +0.207 & -0.132 & -0.203 & -0.055 & 25.mlp & -2.080 & -2.598 & -1.901 & -1.478 & +1.226 & \underline{+0.377} & +1.273 & +2.526 \\
\midrule
\multirow[t]{10}{*}{Qwen 3 32B} & \multirow[t]{5}{*}{Commit} & AIME 2026 & 51.attn & +0.036 & -0.124 & +0.207 & \textbf{-0.246} & \textbf{-0.413} & \textbf{-0.068} & 47.mlp & \underline{-0.461} & \underline{-0.366} & \underline{-0.512} & \underline{-0.551} & \textbf{+0.684} & +0.081 & \textbf{+0.731} & \textbf{+1.536} \\
 &  & GSM8K & 52.attn & -0.052 & -0.192 & +0.097 & \textbf{-0.246} & \underline{-0.412} & \textbf{-0.068} & 48.attn & \textbf{-0.627} & \textbf{-0.609} & \textbf{-0.691} & \textbf{-0.588} & \underline{+0.595} & \textbf{+0.096} & \underline{+0.727} & \underline{+1.202} \\
 &  & OMEGA-500 & 52.mlp & +0.103 & -0.053 & +0.269 & \underline{-0.245} & \underline{-0.412} & \textbf{-0.068} & 46.attn$^{*}$ & -0.185 & -0.082 & -0.156 & -0.369 & +0.413 & +0.073 & +0.544 & +0.785 \\
 &  & GPQA-D & 51.mlp & \textbf{+0.399} & \textbf{+0.186} & \textbf{+0.625} & -0.242 & -0.406 & \textbf{-0.068} & 44.mlp$^{*}$ & -0.100 & -0.050 & -0.042 & -0.237 & +0.223 & +0.053 & +0.294 & +0.404 \\
 &  & MMLU-Pro & 51.mlp & \underline{+0.359} & \underline{+0.152} & \underline{+0.579} & -0.236 & -0.394 & \textbf{-0.068} & 53.mlp$^{*}$ & -0.133 & -0.097 & -0.120 & -0.200 & +0.228 & \underline{+0.086} & +0.441 & +0.217 \\
\cmidrule(lr){2-19}
 & \multirow[t]{5}{*}{Confidence-9} & AIME 2026 & 47.attn & \textbf{+0.423} & \underline{+0.292} & \textbf{+0.563} & \textbf{-0.222} & \underline{-0.374} & \textbf{-0.060} & 51.attn$^{*}$ & -2.632 & -2.705 & -2.844 & -2.301 & \underline{+0.961} & +0.090 & +0.828 & \underline{+2.404} \\
 &  & GSM8K & 48.attn & \underline{+0.385} & +0.276 & \underline{+0.502} & \textbf{-0.222} & \textbf{-0.375} & \underline{-0.059} & 50.mlp & -2.893 & \underline{-3.040} & -3.117 & -2.439 & +0.960 & \textbf{+0.111} & \underline{+0.860} & +2.336 \\
 &  & OMEGA-500 & 48.attn & +0.354 & +0.265 & +0.450 & \underline{-0.211} & -0.355 & -0.057 & 49.attn$^{*}$ & \underline{-3.178} & -2.490 & \textbf{-3.780} & \textbf{-3.576} & +0.902 & +0.045 & +0.746 & +2.348 \\
 &  & GPQA-D & 50.attn & +0.347 & \textbf{+0.295} & +0.402 & -0.108 & -0.178 & -0.034 & 49.attn$^{*}$ & -3.104 & -2.606 & \underline{-3.630} & \underline{-3.296} & +0.887 & +0.057 & +0.714 & +2.312 \\
 &  & MMLU-Pro & 50.attn & +0.306 & +0.260 & +0.355 & -0.123 & -0.194 & -0.047 & 51.attn & \textbf{-3.284} & \textbf{-3.406} & -3.559 & -2.814 & \textbf{+1.002} & \underline{+0.108} & \textbf{+0.862} & \textbf{+2.485} \\
\midrule
\multirow[t]{10}{*}{R1 Qwen 32B} & \multirow[t]{5}{*}{Commit} & AIME 2026 & 47.mlp & \underline{+0.356} & \underline{+0.086} & \underline{+0.646} & -0.374 & -0.581 & \underline{-0.152} & 48.mlp & \textbf{-0.290} & \textbf{-0.165} & \textbf{-0.276} & \textbf{-0.476} & \textbf{+0.273} & \underline{+0.157} & \textbf{+0.288} & \textbf{+0.415} \\
 &  & GSM8K & 48.mlp & -0.053 & -0.247 & +0.155 & -0.377 & -0.585 & \underline{-0.152} & 44.attn$^{\#}$ & -0.071 & -0.060 & -0.052 & -0.108 & +0.094 & +0.079 & +0.091 & +0.117 \\
 &  & OMEGA-500 & 47.mlp & +0.333 & +0.068 & +0.618 & -0.370 & -0.572 & \underline{-0.152} & 44.attn$^{*}$ & -0.079 & -0.066 & -0.045 & \underline{-0.135} & +0.180 & +0.141 & +0.211 & \underline{+0.197} \\
 &  & GPQA-D & 46.mlp & +0.292 & +0.018 & +0.586 & \underline{-0.381} & \underline{-0.593} & \underline{-0.152} & 44.attn$^{*}$ & -0.052 & -0.073 & -0.036 & -0.041 & \underline{+0.201} & \textbf{+0.174} & \underline{+0.252} & +0.180 \\
 &  & MMLU-Pro & 46.attn & \textbf{+0.432} & \textbf{+0.157} & \textbf{+0.728} & \textbf{-0.382} & \textbf{-0.595} & \textbf{-0.153} & 44.attn$^{*}$ & \underline{-0.089} & \underline{-0.102} & \underline{-0.075} & -0.086 & +0.193 & +0.147 & +0.247 & +0.194 \\
\cmidrule(lr){2-19}
 & \multirow[t]{5}{*}{Confidence-9} & AIME 2026 & 46.attn & \underline{+0.344} & +0.262 & \underline{+0.432} & \underline{-0.330} & \underline{-0.505} & \underline{-0.142} & 51.attn & \textbf{-5.198} & \textbf{-5.678} & \textbf{-5.268} & \textbf{-4.464} & +0.891 & +0.288 & +0.760 & +1.861 \\
 &  & GSM8K & 46.attn & \textbf{+0.403} & \textbf{+0.326} & \textbf{+0.485} & \textbf{-0.341} & \textbf{-0.518} & \textbf{-0.150} & 52.attn & -2.331 & -2.346 & -2.344 & -2.294 & \textbf{+1.012} & \textbf{+0.413} & \textbf{+0.868} & \textbf{+1.990} \\
 &  & OMEGA-500 & 46.attn & +0.331 & +0.264 & +0.404 & -0.300 & -0.455 & -0.132 & 52.attn & \underline{-4.403} & \underline{-4.694} & \underline{-4.463} & \underline{-3.937} & +0.889 & +0.299 & +0.762 & +1.839 \\
 &  & GPQA-D & 45.mlp & +0.332 & \underline{+0.270} & +0.398 & -0.266 & -0.389 & -0.133 & 43.attn & -3.888 & -3.774 & -4.088 & -3.817 & \underline{+0.924} & \underline{+0.333} & \underline{+0.790} & \underline{+1.880} \\
 &  & MMLU-Pro & 46.attn & +0.295 & +0.257 & +0.336 & -0.288 & -0.442 & -0.123 & 43.attn & -2.903 & -2.767 & -3.034 & -2.940 & +0.896 & +0.303 & +0.764 & +1.854 \\
\midrule
\bottomrule
\end{tabular}
}
\end{table*}

\begin{table*}[t]
\centering
\small
\setlength{\tabcolsep}{3pt}
\caption{Steering on the test set, broken out by source dataset. See \autoref{tab:steer-test} for details on how to read the table.}
\label{tab:steer-test-by-dataset}
\resizebox{\textwidth}{!}{%
\begin{tabular}{llllrrrrrrrrrrlrrrrrrrrrr}
\toprule
 &  &  & \multicolumn{11}{c}{$\Delta$P(backtrack)} & \multicolumn{11}{c}{$\Delta$E} \\
\cmidrule(lr){4-14} \cmidrule(lr){15-25}
 &  &  & Site & \multicolumn{5}{c}{Negative} & \multicolumn{5}{c}{Positive} & Site & \multicolumn{5}{c}{Negative} & \multicolumn{5}{c}{Positive} \\
\cmidrule(lr){5-9} \cmidrule(lr){10-14} \cmidrule(lr){16-20} \cmidrule(lr){21-25}
Model & Vector & Train set &  & AIME 2026 & GSM8K & OMEGA-500 & GPQA-D & MMLU-Pro & AIME 2026 & GSM8K & OMEGA-500 & GPQA-D & MMLU-Pro &  & AIME 2026 & GSM8K & OMEGA-500 & GPQA-D & MMLU-Pro & AIME 2026 & GSM8K & OMEGA-500 & GPQA-D & MMLU-Pro \\
\midrule
\multirow[t]{10}{*}{Olmo 3 32B} & \multirow[t]{5}{*}{Commit} & AIME 2026 & 35.mlp & \textbf{+0.225} & \underline{+0.161} & \underline{+0.316} & +0.182 & +0.162 & \underline{-0.437} & \textbf{-0.128} & \underline{-0.405} & -0.242 & -0.236 & 20.mlp & \underline{-0.184} & -0.093 & \underline{-0.246} & \underline{-0.352} & \underline{-0.204} & \textbf{+0.784} & \textbf{+0.319} & \textbf{+0.711} & \underline{+0.734} & \textbf{+0.567} \\
 &  & GSM8K & 53.mlp & +0.116 & +0.117 & +0.188 & +0.115 & +0.099 & -0.429 & -0.120 & -0.403 & -0.243 & \underline{-0.243} & 24.attn & \textbf{-0.431} & \textbf{-0.414} & \textbf{-0.461} & \textbf{-0.556} & \textbf{-0.495} & \underline{+0.754} & \underline{+0.241} & \underline{+0.709} & \textbf{+0.788} & \underline{+0.551} \\
 &  & OMEGA-500 & 39.attn & +0.044 & +0.036 & +0.128 & +0.028 & +0.008 & -0.386 & -0.110 & -0.379 & -0.228 & -0.221 & 24.attn$^{*}$ & -0.178 & \underline{-0.122} & -0.169 & -0.270 & -0.193 & +0.327 & +0.155 & +0.267 & +0.339 & +0.257 \\
 &  & GPQA-D & 43.attn & \underline{+0.198} & \textbf{+0.187} & \textbf{+0.326} & \textbf{+0.297} & \textbf{+0.283} & \textbf{-0.456} & \underline{-0.125} & \textbf{-0.409} & \textbf{-0.254} & \textbf{-0.245} & 26.mlp & -0.026 & -0.081 & -0.079 & -0.087 & -0.084 & +0.244 & +0.067 & +0.153 & +0.254 & +0.139 \\
 &  & MMLU-Pro & 43.attn & +0.162 & +0.134 & +0.283 & \underline{+0.227} & \underline{+0.214} & -0.432 & -0.106 & -0.401 & \underline{-0.246} & \textbf{-0.245} & 24.attn$^{*}$ & -0.030 & -0.051 & -0.052 & -0.125 & -0.062 & +0.283 & +0.109 & +0.235 & +0.261 & +0.168 \\
\cmidrule(lr){2-25}
 & \multirow[t]{5}{*}{Confidence-9} & AIME 2026 & 23.attn & \textbf{+0.423} & \textbf{+0.406} & \textbf{+0.464} & \textbf{+0.259} & \textbf{+0.252} & \textbf{-0.332} & \underline{-0.103} & \textbf{-0.325} & \textbf{-0.178} & \underline{-0.184} & 22.attn & -2.621 & -2.141 & -2.726 & -2.005 & -2.058 & +1.143 & +0.471 & +1.445 & +1.675 & +1.078 \\
 &  & GSM8K & 25.attn & \underline{+0.314} & \underline{+0.300} & \underline{+0.351} & \underline{+0.194} & \underline{+0.219} & \underline{-0.301} & \textbf{-0.104} & \underline{-0.289} & \textbf{-0.178} & \textbf{-0.196} & 25.attn & -1.576 & -1.900 & -1.527 & -1.209 & -1.478 & \textbf{+1.175} & \textbf{+0.491} & \underline{+1.497} & \textbf{+1.848} & \textbf{+1.138} \\
 &  & OMEGA-500 & 23.attn & +0.288 & +0.272 & +0.301 & +0.157 & +0.190 & -0.256 & -0.069 & -0.226 & \underline{-0.142} & -0.151 & 23.attn & \underline{-3.383} & \underline{-3.763} & \underline{-3.382} & \underline{-2.768} & \underline{-3.103} & \underline{+1.174} & \underline{+0.490} & \textbf{+1.498} & \underline{+1.844} & \underline{+1.137} \\
 &  & GPQA-D & 25.attn & +0.141 & +0.189 & +0.107 & +0.085 & +0.099 & -0.166 & -0.058 & -0.128 & -0.098 & -0.116 & 23.attn & \textbf{-3.542} & \textbf{-4.089} & \textbf{-3.532} & \textbf{-2.916} & \textbf{-3.433} & +1.167 & +0.486 & +1.488 & +1.809 & +1.127 \\
 &  & MMLU-Pro & 23.attn & +0.233 & +0.233 & +0.256 & +0.127 & +0.156 & -0.196 & -0.054 & -0.164 & -0.117 & -0.128 & 25.mlp & -2.045 & -2.615 & -2.021 & -1.677 & -2.044 & +1.170 & +0.489 & +1.487 & \underline{+1.844} & \underline{+1.137} \\
\midrule
\multirow[t]{10}{*}{Qwen 3 32B} & \multirow[t]{5}{*}{Commit} & AIME 2026 & 51.attn & +0.052 & -0.036 & +0.106 & +0.003 & +0.055 & \textbf{-0.372} & \textbf{-0.161} & \textbf{-0.303} & \textbf{-0.251} & \textbf{-0.142} & 47.mlp & \underline{-0.379} & \underline{-0.505} & \underline{-0.417} & \underline{-0.471} & \underline{-0.534} & \underline{+0.680} & \underline{+0.298} & \underline{+0.888} & \textbf{+0.921} & \textbf{+0.633} \\
 &  & GSM8K & 52.attn & -0.030 & -0.089 & +0.002 & -0.102 & -0.043 & \underline{-0.371} & \textbf{-0.161} & \textbf{-0.303} & \textbf{-0.251} & \textbf{-0.142} & 48.attn & \textbf{-0.527} & \textbf{-0.859} & \textbf{-0.541} & \textbf{-0.542} & \textbf{-0.667} & \textbf{+0.758} & \textbf{+0.345} & \textbf{+0.904} & \underline{+0.740} & +0.226 \\
 &  & OMEGA-500 & 52.mlp & +0.140 & +0.023 & +0.163 & +0.067 & +0.120 & \underline{-0.371} & \textbf{-0.161} & \textbf{-0.303} & \underline{-0.250} & \underline{-0.141} & 46.attn$^{*}$ & -0.176 & -0.139 & -0.188 & -0.215 & -0.207 & +0.440 & +0.265 & +0.523 & +0.475 & \underline{+0.362} \\
 &  & GPQA-D & 51.mlp & \textbf{+0.378} & \textbf{+0.302} & \textbf{+0.409} & \textbf{+0.422} & \textbf{+0.483} & \underline{-0.371} & \underline{-0.156} & \textbf{-0.303} & -0.246 & -0.137 & 44.mlp$^{*}$ & -0.117 & -0.079 & -0.084 & -0.116 & -0.105 & +0.286 & +0.171 & +0.291 & +0.218 & +0.151 \\
 &  & MMLU-Pro & 51.mlp & \underline{+0.357} & \underline{+0.247} & \underline{+0.385} & \underline{+0.380} & \underline{+0.425} & -0.365 & -0.141 & \underline{-0.301} & -0.234 & \underline{-0.141} & 53.mlp$^{*}$ & -0.144 & -0.036 & -0.148 & -0.192 & -0.143 & +0.288 & +0.214 & +0.281 & +0.146 & +0.210 \\
\cmidrule(lr){2-25}
 & \multirow[t]{5}{*}{Confidence-9} & AIME 2026 & 47.attn & \textbf{+0.491} & \textbf{+0.463} & \textbf{+0.449} & \textbf{+0.302} & \textbf{+0.411} & \textbf{-0.349} & \textbf{-0.144} & \underline{-0.270} & \underline{-0.229} & -0.119 & 51.attn$^{*}$ & -2.621 & -2.608 & -2.552 & -2.599 & -2.780 & \underline{+0.895} & +0.420 & \underline{+1.366} & +1.288 & \underline{+0.836} \\
 &  & GSM8K & 48.attn & \underline{+0.434} & \underline{+0.454} & \underline{+0.396} & \underline{+0.286} & \underline{+0.355} & \underline{-0.340} & \underline{-0.137} & \underline{-0.270} & \textbf{-0.231} & \textbf{-0.132} & 50.mlp & -2.940 & \underline{-2.975} & -2.783 & -2.783 & -2.986 & +0.811 & \textbf{+0.483} & +1.322 & \underline{+1.312} & \textbf{+0.873} \\
 &  & OMEGA-500 & 48.attn & +0.417 & +0.412 & +0.350 & +0.251 & +0.342 & -0.333 & -0.130 & \textbf{-0.281} & -0.186 & \underline{-0.124} & 49.attn$^{*}$ & \textbf{-3.415} & -2.689 & \textbf{-3.306} & \underline{-3.230} & \underline{-3.251} & +0.781 & +0.333 & +1.295 & +1.281 & +0.818 \\
 &  & GPQA-D & 50.attn & +0.335 & +0.449 & +0.324 & +0.272 & +0.354 & -0.115 & -0.061 & -0.125 & -0.138 & -0.102 & 49.attn$^{*}$ & \underline{-3.342} & -2.700 & \underline{-3.225} & -3.071 & -3.179 & +0.848 & +0.195 & +1.270 & +1.304 & +0.821 \\
 &  & MMLU-Pro & 50.attn & +0.300 & +0.391 & +0.285 & +0.248 & +0.305 & -0.153 & -0.087 & -0.141 & -0.134 & -0.101 & 51.attn & -3.205 & \textbf{-3.481} & -2.979 & \textbf{-3.246} & \textbf{-3.511} & \textbf{+0.928} & \underline{+0.481} & \textbf{+1.413} & \textbf{+1.313} & \textbf{+0.873} \\
\midrule
\multirow[t]{10}{*}{R1 Qwen 32B} & \multirow[t]{5}{*}{Commit} & AIME 2026 & 47.mlp & \underline{+0.295} & \underline{+0.358} & \underline{+0.341} & \underline{+0.429} & \underline{+0.356} & -0.450 & -0.244 & -0.386 & \underline{-0.430} & -0.363 & 48.mlp & \textbf{-0.292} & \textbf{-0.171} & \textbf{-0.330} & \textbf{-0.377} & \textbf{-0.279} & \textbf{+0.342} & \underline{+0.117} & \textbf{+0.260} & \textbf{+0.359} & \textbf{+0.289} \\
 &  & GSM8K & 48.mlp & -0.135 & +0.015 & -0.074 & -0.026 & -0.045 & \underline{-0.456} & -0.245 & -0.386 & -0.429 & \underline{-0.367} & 44.attn$^{\#}$ & \underline{-0.087} & -0.055 & -0.086 & -0.072 & -0.056 & +0.135 & +0.057 & +0.112 & +0.090 & +0.075 \\
 &  & OMEGA-500 & 47.mlp & +0.259 & +0.357 & +0.313 & +0.403 & +0.333 & -0.446 & -0.237 & -0.383 & -0.428 & -0.353 & 44.attn$^{*}$ & -0.052 & -0.081 & \underline{-0.087} & -0.102 & -0.071 & +0.239 & +0.078 & +0.189 & +0.218 & +0.175 \\
 &  & GPQA-D & 46.mlp & +0.215 & +0.317 & +0.275 & +0.353 & +0.300 & \textbf{-0.460} & \underline{-0.247} & \underline{-0.390} & \textbf{-0.434} & \textbf{-0.372} & 44.attn$^{*}$ & -0.002 & -0.078 & -0.052 & -0.067 & -0.060 & \underline{+0.249} & \textbf{+0.118} & \underline{+0.217} & +0.233 & +0.188 \\
 &  & MMLU-Pro & 46.attn & \textbf{+0.357} & \textbf{+0.491} & \textbf{+0.409} & \textbf{+0.468} & \textbf{+0.435} & \textbf{-0.460} & \textbf{-0.248} & \textbf{-0.394} & \textbf{-0.434} & \textbf{-0.372} & 44.attn$^{*}$ & -0.027 & \underline{-0.110} & -0.085 & \underline{-0.117} & \underline{-0.106} & +0.230 & +0.096 & +0.196 & \underline{+0.248} & \underline{+0.194} \\
\cmidrule(lr){2-25}
 & \multirow[t]{5}{*}{Confidence-9} & AIME 2026 & 46.attn & \underline{+0.382} & \underline{+0.390} & \underline{+0.412} & \underline{+0.280} & \underline{+0.256} & \underline{-0.397} & \underline{-0.212} & \textbf{-0.342} & \underline{-0.377} & \underline{-0.324} & 51.attn & \textbf{-4.836} & \textbf{-5.704} & \textbf{-5.213} & \textbf{-4.922} & \textbf{-5.315} & +1.270 & +0.313 & +0.911 & \underline{+1.204} & \underline{+0.757} \\
 &  & GSM8K & 46.attn & \textbf{+0.418} & \textbf{+0.525} & \textbf{+0.461} & \textbf{+0.310} & \textbf{+0.300} & \textbf{-0.416} & \textbf{-0.218} & \underline{-0.339} & \textbf{-0.391} & \textbf{-0.341} & 52.attn & -2.239 & -2.440 & -2.395 & -2.254 & -2.325 & \textbf{+1.410} & \textbf{+0.460} & \textbf{+1.071} & \textbf{+1.279} & \textbf{+0.838} \\
 &  & OMEGA-500 & 46.attn & +0.353 & +0.379 & +0.399 & +0.274 & +0.252 & -0.360 & -0.192 & -0.310 & -0.336 & -0.299 & 52.attn & \underline{-4.141} & \underline{-4.717} & \underline{-4.466} & \underline{-4.222} & \underline{-4.467} & +1.274 & +0.315 & +0.919 & +1.193 & +0.747 \\
 &  & GPQA-D & 45.mlp & +0.366 & +0.374 & +0.393 & +0.275 & +0.251 & -0.318 & -0.176 & -0.276 & -0.292 & -0.267 & 43.attn & -3.784 & -3.711 & -3.999 & -3.897 & -4.048 & \underline{+1.320} & \underline{+0.369} & \underline{+0.988} & +1.185 & +0.755 \\
 &  & MMLU-Pro & 46.attn & +0.335 & +0.317 & +0.359 & +0.244 & +0.221 & -0.352 & -0.178 & -0.302 & -0.323 & -0.286 & 43.attn & -2.803 & -2.773 & -3.016 & -2.918 & -3.007 & +1.286 & +0.333 & +0.952 & +1.172 & +0.736 \\
\midrule
\bottomrule
\end{tabular}
}
\end{table*}

\FloatBarrier

\section{Confidence steering on linguistic confidence}
Additional examples of how steering affects the language of the model are shown in \autoref{tab:steering_qualitative_qwen3_32b} and \autoref{tab:steering_qualitative_r1qwen_32b}. See \autoref{sec:5steer_qualitative} for discussion.

\begin{table}[htbp]
  \centering
  \footnotesize
  \caption{Effect of regenerating an answer proposal with $\alpha=-0.5$ confidence steering on Qwen 3 32B (GPQA-D, problem index 47). Both continuations share the same prefix. Removing confidence causes the model to revisit other options.}
  \label{tab:steering_qualitative_qwen3_32b}
  \begin{tabular}{@{}>{\raggedright\arraybackslash}p{0.47\linewidth}>{\raggedright\arraybackslash}p{0.47\linewidth}@{}}
    \toprule
    \multicolumn{2}{@{}>{\raggedright\arraybackslash}p{0.96\linewidth}@{}}{%
      \textbf{Problem.} Cyclohexanone was treated with bromine, forming product 1.
      1 was heated with sodium hydroxide, forming product 2.
      2 was treated with thionyl chloride and pyridine, forming 3.
      3 was treated with lithium tri-\emph{tert}-butoxyaluminum hydride, forming product 4.
      How many chemically distinct hydrogen atoms are there on product 4? \textbf{Ground Truth.} D.
      \newline
      (A) 10 \quad (B) 7 \quad (C) 8 \quad (D) 6} \\
    \midrule
    \multicolumn{2}{@{}>{\raggedright\arraybackslash}p{0.96\linewidth}@{}}{%
      \textbf{Shared prefix (through marker 3).}
      \itshape But let me verify once more. If product 4 is cyclopentanol, with 6 distinct hydrogen environments, then answer D (6) is correct.} \\
    \midrule
    \textbf{Unsteered} & \textbf{Steered} \\
    \midrule
    \itshape Therefore, the answer is D. \newline
    \texttt{</think>} \newline
    To determine the\dots
    &
    \itshape But let's think about another possibility: maybe product 4 is not cyclopentanol. Let me re-examine the steps. \newline\newline
    Product 2: cyclopentanecarboxylic acid? Or something else?\dots \\
    \midrule
    Commits to D and exits the reasoning block.
    & Reopens the question and revisits an earlier intermediate. \\
    \bottomrule
  \end{tabular}
\end{table}

\begin{table}[htbp]
  \centering
  \footnotesize
  \caption{Effect of regenerating an answer proposal with $\alpha=-0.5$ confidence steering on Distill 32B (GSM8K, problem index 785). Both continuations share the same prefix. Negative steering causes the model to re-checks the problem statement and reopen computation.}
  \label{tab:steering_qualitative_r1qwen_32b}
  \begin{tabular}{@{}>{\raggedright\arraybackslash}p{0.47\linewidth}>{\raggedright\arraybackslash}p{0.47\linewidth}@{}}
    \toprule
    \multicolumn{2}{@{}>{\raggedright\arraybackslash}p{0.96\linewidth}@{}}{%
      \textbf{Problem.} Carrie was given ten twenties and 140 quarters by her aunt
      to use for lunch. If she spent all the quarters and $3/5$ of the twenties,
      calculate the total amount of money she paid for the lunch. \textbf{Ground Truth.} 155.} \\
    \midrule
    \multicolumn{2}{@{}>{\raggedright\arraybackslash}p{0.96\linewidth}@{}}{%
      \textbf{Shared prefix (through marker 4).}
      \itshape Adding those together, 35 + 120 is 155. Yep, that seems accurate.} \\
    \midrule
    \textbf{Unsteered} & \textbf{Steered} \\
    \midrule
    \itshape So, I think I'm confident that Carrie spent a total of \$155 on lunch. \newline\newline
    \textbf{Final Answer} \newline
    The total amount Carrie paid for lunch is $\boxed{155}$ dollars. \newline
    \texttt{</think>} \newline
    Carrie was given ten twenties and 140 quarters by her aunt. To find the total amount\dots
    &
    \itshape So, I think the total amount Carrie paid for lunch is \$155. Let me just make sure I didn't miss anything in the problem. She was given ten twenties and 140 quarters. She spent all the quarters and $3/5$ of the twenties. So, I think I accounted for that correctly. \newline\newline
    Wait, hold on, let me think again\dots \\
    \midrule
    States the answer and exits the reasoning block.
    & Re-reads the problem, then reopens the computation. \\
    \bottomrule
  \end{tabular}
\end{table}
\FloatBarrier

\section{Multi Token Steering Details}
\subsection{CoT Manipulation vs Steering Details}
\label{app:multitoken_misc}

\autoref{tab:cot_vs_steer_sentences} reports the confidence, uncertain, and neutral sentences used in \cref{sec:5cot_vs_steer}.
\label{app:cot_vs_steering_misc}
\autoref{tab:cot_vs_steer_controls} records the baseline \texttt{</think>}\,/\,$p_{\mathrm{bt}}$ used in \autoref{tab:cot_vs_steer_primary_append}.

\begin{table}[htbp]
    \centering
    \caption{Confident, uncertain, and neutral sentences used in the surface-level vs latent confidence experiments. Let \texttt{X} be the answer value. We do not replace the final punctuation of the sentence, but rather use the same punctuation token that the original answer proposal sentence uses.}
    \label{tab:cot_vs_steer_sentences}
    \begin{tabular}{l|l}
        Sentence Type & Sentence \\
        \toprule
        Confident Sentence 1 & \texttt{Therefore, I am confident the answer is X} \\
        Confident Sentence 2 & \texttt{Therefore, my final answer is X} \\
        Confident Sentence 3 & \texttt{Therefore, the answer is \textbackslash boxed \{X\}} \\
        \midrule
        Uncertain Sentence 1 & \texttt{I am not confident, but I think the answer is X} \\
        Uncertain Sentence 2 & \texttt{Therefore, the answer may be X} \\
        Uncertain Sentence 3 & \texttt{Perhaps the answer is X} \\
        \midrule
        Neutral Sentence 1 & \texttt{So the answer is X} \\
        \bottomrule
    \end{tabular}
\end{table}

\begin{table}[htbp]
  \centering
  \scriptsize
  \setlength{\tabcolsep}{4pt}
  \caption{\texttt{</think>}\,/\,$p_{\mathrm{bt}}$ baselines for surface-level vs latent confidence results in \autoref{tab:cot_vs_steer_primary_append}. 
    The $\Delta$s in \autoref{tab:cot_vs_steer_primary_append} are computed relative t each method's own baseline values.
    The slight numerical variations are due to bfloat16 vLLM implementation differences. $n$ is the total number of rollouts generated (four per marker) from the markers that survive filtering.}
  \label{tab:cot_vs_steer_controls}
  \begin{tabular}{llcccc||cccc}
    \toprule
    \multirow{2}{*}{Model} & \multirow{2}{*}{Metric} & \multicolumn{4}{c||}{Positive: confident proposal (mid-trace markers)} & \multicolumn{4}{c}{Negative: hedged proposal (commitment markers)} \\
    \cmidrule(lr){3-6} \cmidrule(lr){7-10}
     &  & transplant & append & steer ($\alpha{=}0$) & \multicolumn{1}{c||}{$n$} & transplant & append & steer ($\alpha{=}0$) & $n$ \\
    \midrule
    \multirow{2}{*}{OLMo-3-32B} & \texttt{</think>} & 3.2 & 3.5 & 3.5 & \multirow{2}{*}{564} & 97.2 & 97.1 & 97.3 & \multirow{2}{*}{792} \\
     & $p_{\mathrm{bt}}$ & 84.8 & 84.8 & 85.9 &  & 4.9 & 4.9 & 5.1 &  \\
    \midrule
    \multirow{2}{*}{Qwen3-32B} & \texttt{</think>} & 6.8 & 6.8 & 6.8 & \multirow{2}{*}{456} & 97.7 & 97.7 & 97.7 & \multirow{2}{*}{684} \\
     & $p_{\mathrm{bt}}$ & 82.9 & 82.9 & 83.0 &  & 5.4 & 5.4 & 5.9 &  \\
    \midrule
    \multirow{2}{*}{R1-Qwen-32B} & \texttt{</think>} & 11.2 & 11.4 & 11.1 & \multirow{2}{*}{668} & 91.7 & 91.7 & 91.7 & \multirow{2}{*}{544} \\
     & $p_{\mathrm{bt}}$ & 82.8 & 82.8 & 82.9 &  & 13.9 & 13.9 & 13.9 &  \\
    \bottomrule
  \end{tabular}
\end{table}

\FloatBarrier

\subsection{Post Marker Steering Details}
\label{app:postmarkersteer}
\autoref{tab:post-marker-steer-compact} shows the full results of steering after answer proposal markers in the pooled test set, as described in \cref{sec:5postmarker}. 

\revise{\paragraph{Prompts with 2+ unique answers} As shown in \autoref{tab:unique-vs-proposed-summary}, 11-16\% of rollouts propose more than one unique answer.
To understand how steering impacts prompts whose rollouts may contain two or more unique answers, we operationalize this through two scenarios:

\begin{enumerate}
    \item Among incorrect answer proposals that the model can demonstrably recover from (the current answer is incorrect but the model has proposed a correct answer in either this rollout or any other rollout that shares the same prompt), how often does steering lead to the correct answer?
    \item Among correct answer proposals where the model can demonstrably arrive at the wrong answer (the current proposal is correct but the model has committed to an incorrect answer in either this rollout or any other rollout that shares the same prompt), how often does steering lead to the incorrect answer?
\end{enumerate}

We obtain relevant proposal markers from the test set by checking the proposals of all rollouts that share the same prompt as the relevant marker. We analyze the results using the steered continuations with the best positive and negative $\alpha$ per model, as well as the $\alpha=0$ control (\autoref{sec:5postmarker}).

Results for scenario 1 are in \autoref{tab:recovery_incorrect}.  
Against the control, positive steering reduces the model's ability to recover the correct answer, confirming a small cost to suppressing backtracking. 
Conversely, negative steering slightly increases the model’s ability to recover the correct answer, likely by reducing the model's confidence on its incorrect answer and leading to re-evaluation.

Results for scenario 2 in \autoref{tab:retention_correct}. Positive steering is comparable to or slightly better than no steering at maintaining the correct answer. Combined with scenario 1, positive steering has a small net effect, consistent with the near zero accuracy change in \autoref{tab:post-marker-steer-compact}.}


\begin{table}[t]
\centering
\caption{\revise{Scenario 1: recovery from incorrect answer proposals. Markers whose proposal is incorrect, where another rollout on the same prompt reached the correct answer. I$\rightarrow$C counts continuations that end with the correct answer; ``Same answer'' counts continuations whose final answer matches the proposal.}}
\label{tab:recovery_incorrect}
\begin{tabular}{llrrrr}
\toprule
Model & $\alpha$ & Markers & Conts & I$\rightarrow$C & Same answer \\
\midrule
\multirow{3}{*}{Olmo 3 32B}  & $0$     & 65 & 260 & 42 (16\%)          & 207 (80\%) \\
& $-0.25$ & 65 & 260 & \textbf{48 (18\%)} & 191 (73\%) \\
& $+0.75$ & 65 & 260 & 28 (11\%)          & 226 (87\%) \\
\midrule
\multirow{3}{*}{Qwen 3 32B}  & $0$     & 57 & 228 & 36 (16\%)          & 176 (77\%) \\
& $-0.25$ & 57 & 228 & \textbf{40 (18\%)} & 166 (73\%) \\
& $+0.25$ & 57 & 228 & 34 (15\%)          & 173 (76\%) \\
\midrule
\multirow{3}{*}{Distill 32B} & $0$     & 64 & 256 & 43 (17\%)          & 185 (72\%) \\
& $-0.1$  & 64 & 256 & \textbf{48 (19\%)} & 180 (70\%) \\
& $+0.5$  & 64 & 256 & 35 (14\%)          & 205 (80\%) \\
\bottomrule
\end{tabular}
\end{table}

\begin{table}[t]
\centering
\caption{\revise{Scenario 2: retention of correct answer proposals. Markers whose proposal is correct, where another rollout on the same prompt ended with an incorrect answer. C$\rightarrow$I counts continuations that end with an incorrect answer.}}
\label{tab:retention_correct}
\begin{tabular}{llrrrr}
\toprule
Model & $\alpha$ & Markers & Conts & C$\rightarrow$I & Stays correct \\
\midrule
\multirow{3}{*}{Olmo 3 32B}  & $0$     & 68 & 272 & 20 (7\%)  & 252 (93\%)          \\
& $-0.25$ & 68 & 272 & 24 (9\%)  & 248 (91\%)          \\
& $+0.75$ & 68 & 272 & 11 (4\%)  & \textbf{261 (96\%)} \\
\midrule
\multirow{3}{*}{Qwen 3 32B}  & $0$     & 40 & 160 & 19 (12\%) & \textbf{141 (88\%)} \\
& $-0.25$ & 40 & 160 & 21 (13\%) & 139 (87\%)          \\
& $+0.25$ & 40 & 160 & 20 (12\%) & \textbf{140 (88\%)} \\
\midrule
\multirow{3}{*}{Distill 32B} & $0$     & 55 & 220 & 19 (9\%)  & 201 (91\%)          \\
& $-0.1$  & 55 & 220 & 28 (13\%) & 192 (87\%)          \\
& $+0.5$  & 55 & 220 & 14 (6\%)  & \textbf{206 (94\%)} \\
\bottomrule
\end{tabular}
\end{table}

\begin{table}[htbp]
  \centering
  \scriptsize
  \setlength{\tabcolsep}{3pt}
  \caption{Post-marker continuation steering on the test set. Every steered continuation is compared with the unsteered ($\alpha=0$) continuation resampled from the same prefix with the same seed. The $\alpha=0$ rows compares against the stored rollout the marker was taken from. $\Delta L$ is the mean difference in generated length; C$\to$I and I$\to$C count answers that flipped between incorrect and correct, and $\Delta$acc is the net change in accuracy. $\Delta$prop is the mean change in the number of judged answer proposals after the marker. short is the fraction of continuations shorter than their baseline, averaged per marker. The superscript on $\alpha$ is the number of steered continuations sampled per marker. The standard error of the mean is in the subscript.}
  \label{tab:post-marker-steer-compact}
\resizebox{\textwidth}{!}{
  \begin{tabular}{lllccccccrrrr}
    \toprule
    Model & Vec & $\alpha$ & \multicolumn{3}{c}{$\overline{\Delta L}$} & \multicolumn{3}{c}{$\Delta$prop} & short & C$\to$I & I$\to$C & $\Delta$acc \\
    \cmidrule(lr){4-6} \cmidrule(lr){7-9}
     &  &  & all & mid-trace & final & all & mid-trace & final &  &  &  &  \\
    \midrule
    \multirow{15}{*}{OLMo-3-32B} & none & $0^{4}$ & $-25$ & $-81_{\pm 77}$ & $36_{\pm 12}$ & $-0.37$ & $-0.97_{\pm 0.37}$ & $+0.26_{\pm 0.04}$ & 0.49 & 7.75 & 2.25 & -1.1 \\
    \cmidrule(l){2-13}
     & \multirow{7}{*}{commit} & $-0.5^{4}$ & $927$ & $1342_{\pm 150}$ & $482_{\pm 74}$ & $+6.16$ & $+9.50_{\pm 1.23}$ & $+2.60_{\pm 0.43}$ & 0.28 & 7.75 & 7.25 & -0.1 \\
     &  & $-0.25^{4}$ & $343$ & $516_{\pm 66}$ & $156_{\pm 49}$ & -- & -- & -- & 0.39 & 6.50 & 6.50 & +0.0 \\
     &  & $-0.1^{4}$ & $109$ & $167_{\pm 61}$ & $46_{\pm 14}$ & -- & -- & -- & 0.42 & 7.25 & 7.25 & +0.0 \\
     &  & $0.1^{4}$ & $-80$ & $-140_{\pm 52}$ & $-15_{\pm 9}$ & -- & -- & -- & 0.47 & 4.75 & 6.25 & +0.3 \\
     &  & $0.25^{4}$ & $-334$ & $-695_{\pm 87}$ & $53_{\pm 48}$ & $-2.34$ & $-4.44_{\pm 0.48}$ & $-0.09_{\pm 0.04}$ & 0.63 & 9.75 & 7 & -0.5 \\
     &  & $0.5^{1}$ & $3483$ & $3744_{\pm 737}$ & $3203_{\pm 499}$ & -- & -- & -- & 0.58 & 85 & 5 & -16.0 \\
     &  & $0.75^{1}$ & $20524$ & $20681_{\pm 601}$ & $20356_{\pm 624}$ & -- & -- & -- & 0.05 & 341 & 0 & -68.2 \\
    \cmidrule(l){2-13}
     & \multirow{7}{*}{conf} & $-0.5^{1}$ & $8151$ & $11672_{\pm 441}$ & $4366_{\pm 372}$ & -- & -- & -- & 0.07 & 34 & 8 & -5.2 \\
     &  & $-0.25^{4}$ & $2101$ & $3472_{\pm 173}$ & $627_{\pm 87}$ & $+3.73$ & $+6.01_{\pm 0.54}$ & $+1.30_{\pm 0.28}$ & 0.13 & 6.25 & 7.25 & +0.2 \\
     &  & $-0.1^{4}$ & $611$ & $1018_{\pm 70}$ & $175_{\pm 45}$ & -- & -- & -- & 0.28 & 5.25 & 7.25 & +0.4 \\
     &  & $0.1^{4}$ & $-356$ & $-647_{\pm 66}$ & $-43_{\pm 10}$ & -- & -- & -- & 0.62 & 7.50 & 6.50 & -0.2 \\
     &  & $0.25^{4}$ & $-694$ & $-1282_{\pm 92}$ & $-62_{\pm 11}$ & -- & -- & -- & 0.78 & 6.50 & 7.25 & +0.1 \\
     &  & $0.5^{4}$ & $-953$ & $-1742_{\pm 115}$ & $-105_{\pm 13}$ & -- & -- & -- & 0.85 & 7.25 & 7.25 & +0.0 \\
     &  & $0.75^{4}$ & $-1057$ & $-1906_{\pm 121}$ & $-144_{\pm 15}$ & $-3.29$ & $-6.21_{\pm 0.53}$ & $-0.16_{\pm 0.04}$ & 0.91 & 10.25 & 7 & -0.6 \\
    \midrule
    \multirow{15}{*}{Qwen3-32B} & none & $0^{4}$ & $-29$ & $-84_{\pm 84}$ & $30_{\pm 10}$ & $-0.36$ & $-0.89_{\pm 0.31}$ & $+0.21_{\pm 0.04}$ & 0.50 & 3 & 7 & +0.8 \\
    \cmidrule(l){2-13}
     & \multirow{7}{*}{commit} & $-0.5^{1}$ & $22967$ & $22980_{\pm 462}$ & $22954_{\pm 486}$ & -- & -- & -- & 0.00 & 74 & 3 & -18.9 \\
     &  & $-0.25^{4}$ & $532$ & $797_{\pm 80}$ & $249_{\pm 26}$ & $+4.11$ & $+6.36_{\pm 0.59}$ & $+1.70_{\pm 0.16}$ & 0.28 & 8 & 6.50 & -0.3 \\
     &  & $-0.1^{4}$ & $186$ & $311_{\pm 72}$ & $52_{\pm 12}$ & -- & -- & -- & 0.42 & 5.50 & 5.25 & -0.1 \\
     &  & $0.1^{4}$ & $-92$ & $-158_{\pm 45}$ & $-21_{\pm 6}$ & -- & -- & -- & 0.56 & 5.25 & 5.50 & +0.0 \\
     &  & $0.25^{4}$ & $-198$ & $-386_{\pm 48}$ & $2_{\pm 32}$ & $-1.34$ & $-2.42_{\pm 0.31}$ & $-0.19_{\pm 0.03}$ & 0.61 & 6.75 & 6 & -0.2 \\
     &  & $0.5^{1}$ & $-714$ & $-1109_{\pm 195}$ & $-293_{\pm 30}$ & -- & -- & -- & 0.82 & 102 & 2 & -20.0 \\
     &  & $0.75^{1}$ & $1953$ & $1768_{\pm 621}$ & $2150_{\pm 528}$ & -- & -- & -- & 0.83 & 323 & 2 & -64.2 \\
    \cmidrule(l){2-13}
     & \multirow{7}{*}{conf} & $-0.5^{1}$ & $14380$ & $19624_{\pm 706}$ & $8789_{\pm 765}$ & -- & -- & -- & 0.14 & 209 & 3 & -41.2 \\
     &  & $-0.25^{4}$ & $1250$ & $2183_{\pm 192}$ & $254_{\pm 36}$ & $+3.04$ & $+4.93_{\pm 0.39}$ & $+1.01_{\pm 0.14}$ & 0.27 & 11 & 7.50 & -0.7 \\
     &  & $-0.1^{4}$ & $335$ & $602_{\pm 65}$ & $51_{\pm 21}$ & -- & -- & -- & 0.38 & 5.50 & 5.50 & +0.0 \\
     &  & $0.1^{4}$ & $-148$ & $-284_{\pm 45}$ & $-2_{\pm 7}$ & -- & -- & -- & 0.58 & 6 & 5.75 & -0.1 \\
     &  & $0.25^{4}$ & $-313$ & $-618_{\pm 73}$ & $11_{\pm 8}$ & $-1.08$ & $-1.98_{\pm 0.24}$ & $-0.10_{\pm 0.03}$ & 0.61 & 7.50 & 4.50 & -0.6 \\
     &  & $0.5^{1}$ & $-223$ & $-473_{\pm 293}$ & $45_{\pm 62}$ & -- & -- & -- & 0.66 & 23 & 7 & -3.2 \\
     &  & $0.75^{1}$ & $1380$ & $2170_{\pm 559}$ & $538_{\pm 244}$ & -- & -- & -- & 0.64 & 76 & 8 & -13.6 \\
    \midrule
    \multirow{13}{*}{R1-Qwen-32B} & none & $0^{4}$ & $346$ & $537_{\pm 184}$ & $141_{\pm 55}$ & $+0.40$ & $+0.38_{\pm 0.44}$ & $+0.42_{\pm 0.06}$ & 0.47 & 8.25 & 5.75 & -0.5 \\
    \cmidrule(l){2-13}
     & \multirow{6}{*}{commit} & $-0.25^{1}$ & $9118$ & $12785_{\pm 862}$ & $5177_{\pm 620}$ & -- & -- & -- & 0.11 & 91 & 6 & -17.0 \\
     &  & $-0.1^{4}$ & $2021$ & $3193_{\pm 439}$ & $761_{\pm 176}$ & $+3.01$ & $+4.73_{\pm 0.57}$ & $+1.25_{\pm 0.20}$ & 0.29 & 21.50 & 7 & -2.9 \\
     &  & $0.1^{4}$ & $-492$ & $-831_{\pm 141}$ & $-127_{\pm 54}$ & -- & -- & -- & 0.58 & 7 & 9 & +0.4 \\
     &  & $0.25^{4}$ & $-733$ & $-1285_{\pm 184}$ & $-139_{\pm 55}$ & -- & -- & -- & 0.70 & 6.25 & 8.25 & +0.4 \\
     &  & $0.5^{4}$ & $-880$ & $-1657_{\pm 196}$ & $-44_{\pm 94}$ & $-3.49$ & $-6.35_{\pm 0.52}$ & $-0.41_{\pm 0.06}$ & 0.77 & 9.75 & 7.75 & -0.4 \\
     &  & $0.75^{1}$ & $2052$ & $1416_{\pm 697}$ & $2736_{\pm 594}$ & -- & -- & -- & 0.72 & 64 & 11 & -10.6 \\
    \cmidrule(l){2-13}
     & \multirow{6}{*}{conf} & $-0.25^{1}$ & $5699$ & $7740_{\pm 671}$ & $3507_{\pm 463}$ & -- & -- & -- & 0.12 & 48 & 13 & -7.0 \\
     &  & $-0.1^{4}$ & $1082$ & $1877_{\pm 245}$ & $228_{\pm 46}$ & $+1.99$ & $+3.31_{\pm 0.59}$ & $+0.60_{\pm 0.13}$ & 0.29 & 14 & 8 & -1.2 \\
     &  & $0.1^{4}$ & $-440$ & $-751_{\pm 136}$ & $-105_{\pm 36}$ & -- & -- & -- & 0.62 & 6.75 & 7 & +0.1 \\
     &  & $0.25^{4}$ & $-722$ & $-1239_{\pm 168}$ & $-167_{\pm 55}$ & -- & -- & -- & 0.75 & 7 & 8.75 & +0.4 \\
     &  & $0.5^{4}$ & $-839$ & $-1446_{\pm 194}$ & $-188_{\pm 55}$ & $-2.41$ & $-4.36_{\pm 0.45}$ & $-0.31_{\pm 0.05}$ & 0.79 & 9 & 8.75 & -0.0 \\
     &  & $0.75^{1}$ & $-317$ & $-694_{\pm 517}$ & $89_{\pm 271}$ & -- & -- & -- & 0.80 & 17 & 8 & -1.8 \\
    \bottomrule
  \end{tabular}
}
\end{table}

\FloatBarrier

\subsection{Steering at Inference Details}
\begin{table}[htbp]
    \centering
    \scriptsize
    \caption{\revise{95\% bootstrap confidence intervals for the results in \autoref{tab:steer-scratch} on confidence steering at inference time. The accuracy delta and response length delta are taken with respect to the unsteered rollouts. All steered rollouts use $\alpha=0.1$.}}
    \setlength{\tabcolsep}{2pt}
    \label{tab:steer-scratch-cis}
\resizebox{\textwidth}{!}{
\begin{tabular}{ll|cc|cc|cc|cc|cc|cc}
\toprule
Model & Arm & \multicolumn{2}{c}{AIME 2026} & \multicolumn{2}{c}{GSM8K} & \multicolumn{2}{c}{OMEGA-500} & \multicolumn{2}{c}{GPQA-D} & \multicolumn{2}{c}{MMLU-Pro} & \multicolumn{2}{c}{Avg} \\
\cmidrule(lr){3-4}\cmidrule(lr){5-6}\cmidrule(lr){7-8}\cmidrule(lr){9-10}\cmidrule(lr){11-12}\cmidrule(lr){13-14}
 & & $\Delta$Acc & $\Delta$Len & $\Delta$Acc & $\Delta$Len & $\Delta$Acc & $\Delta$Len & $\Delta$Acc & $\Delta$Len & $\Delta$Acc & $\Delta$Len & $\Delta$Acc & $\Delta$Len \\
\midrule
\multirow{2}{*}{Olmo 3 32B} & $d$=0 & [-13.3, +2.5] & [-17.3, +0.3]\% & [-1.5, +0.1] & [-24.3, -19.6]\% & [-5.2, +0.0] & [-13.7, -8.9]\% & [-6.2, +0.9] & [-16.1, -10.8]\% & [-3.1, +1.6] & [-23.8, -16.4]\% & [-4.3, -0.5] & [-17.2, -13.0]\% \\
 & $d$=5000 & [-6.7, +3.3] & [-11.8, +0.1]\% & [-0.9, +0.5] & [-4.6, +2.0]\% & [-4.6, +0.6] & [-10.6, -6.6]\% & [-2.4, +3.7] &[-12.9, -8.1]\% & [-2.9, +1.8] & [-11.9, -5.7]\% & [-2.1, +0.6] & [-8.6, -5.4]\% \\
\midrule
\multirow{2}{*}{Qwen 3 32B} & $d$=0 & [-9.2, +6.7] & [-15.8, +2.1]\% & [-2.2, +0.4] & [-24.1, -12.5]\% & [-2.1, +3.9] & [-8.2, -2.4]\% & [-2.7, +3.4] & [-9.4, -2.7]\% & [-2.8, +0.9] & [-9.4, -3.1]\% & [-2.2, +1.5] & [-10.9, -6.1]\% \\
 & $d$=5000 & [-7.5, +6.7] & [-13.3, -2.3]\% & [-0.8, +0.4] & [-1.4, +4.2]\% & [-1.0, +4.8] & [-8.7, -3.9]\% & [-2.1, +3.4] & [-3.5, +1.9]\% & [-2.0, +2.4] & [-4.8, +1.9]\% & [-1.2, +2.1] & [-4.6, -1.4]\% \\
\midrule
\multirow{2}{*}{Distill 32B} & $d$=0 & [-9.2, +3.3] & [-25.9, -10.2]\% & [-0.8, +1.6] & [-33.1, -18.8]\% & [-7.8, -2.1] & [-22.0, -14.3]\% & [-4.4, +1.9] & [-23.0, -14.7]\% & [-0.4, +4.9] & [-26.8, -10.0]\% & [-2.9, +0.4] & [-22.9, -17.0]\% \\
 & $d$=5000 & [-13.3, +3.3] & [-21.0, +1.8]\% & [-0.6, +1.1] & [-16.8, -0.9]\% & [-3.8, +2.0] & [-18.2, -10.3]\% & [-1.4, +5.2] & [-13.3, -4.7]\% & [-1.9, +2.6] & [-16.6, -1.4]\% & [-2.7, +1.3] & [-13.5, -6.8]\% \\
\bottomrule
\end{tabular}}
\end{table}
\FloatBarrier

\section{Full Logit Lens}
\label{app:logitlens}
We provide the unfiltered logit lens tables for $v_\text{conf}$ and $v_\text{commit}$ in \autoref{tab:logit_lens_vconf} and \autoref{tab:logit_lens_vcommit} respectively. See the discussion in \cref{sec:4logit-lens}.

\newcommand{\lbsize}{\fontsize{7}{8.4}\selectfont} 

\begin{table*}[htbp]
\centering
\footnotesize
\setlength{\tabcolsep}{4pt}
\renewcommand{\arraystretch}{1.15}
\caption{Unfiltered top-10 logit lens tokens for confidence direction $v_\text{conf}$, sorted in descending order by logit value.}
\label{tab:logit_lens_vconf}
\begin{tabular}{@{}>{\lbsize}l >{\lbsize}l >{\raggedright\arraybackslash\ttfamily\scriptsize}p{0.85\linewidth}@{}}
\toprule
Model & Direction & Top-10 logit lens tokens \\
\midrule
\multirow{2}{*}{Olmo 3 32B}
& Negative & ` somewhat', ` questionable', ` controversial', ` slightly', ` vaguely', ` uncertain', ` unclear', `-ish', ` unfortunately', `=False' \\
\addlinespace
& Positive & ` strongest', `ses', `zes', ` perfected', ` superior', `[U+FFFD]', ` stronger', ` flawless', ` perfectly', `/all' \\
\midrule
\multirow{2}{*}{Qwen 3 32B}
& Negative & \begin{CJK*}{UTF8}{gbsn}`有一定的', `赦', `具有一定', `部分内容', `悬', `鹄', `幕后', `有一定', `主要领导', `好吗'\end{CJK*} \\
\addlinespace
& Positive & `ourt', \begin{CJK*}{UTF8}{gbsn}`无疑是', `完全可以'\end{CJK*}, `ock', \begin{CJK*}{UTF8}{gbsn}`毫无疑', `完美的'\end{CJK*}, `\textcyrillic{ следует}', \begin{CJK*}{UTF8}{gbsn}`享有', `赞', `做强'\end{CJK*} \\
\midrule
\multirow{2}{*}{Distill 32B}
& Negative & \begin{CJK*}{UTF8}{gbsn}`不一定', `不确定', `不是很', `勉强', `未必'\end{CJK*}, ` uncertain', \begin{CJK*}{UTF8}{gbsn}`一半', `初步', `不如', `有一定的'\end{CJK*} \\
\addlinespace
& Positive & \begin{CJK*}{UTF8}{gbsn}`放心'\end{CJK*}, ` flawless', ` redund', `Trivia', `\'{\u a}ng', \begin{CJK*}{UTF8}{gbsn}`康'\end{CJK*}, `ubern', \begin{CJK*}{UTF8}{gbsn}`无疑', `久', `亲自'\end{CJK*} \\
\bottomrule
\end{tabular}
\end{table*}

\begin{table*}[htbp]
\centering
\footnotesize
\setlength{\tabcolsep}{4pt}
\renewcommand{\arraystretch}{1.15}
\caption{\revise{Unfiltered top-10 logit lens tokens for commitment direction $v_\text{commit}$, sorted in descending order by logit value.}}
\label{tab:logit_lens_vcommit}
\begin{tabular}{@{}>{\lbsize}l >{\lbsize}l >{\raggedright\arraybackslash\ttfamily\scriptsize}p{0.85\linewidth}@{}}
\toprule
Model & Direction & Top-10 logit lens tokens \\
\midrule
\multirow{2}{*}{Olmo 3 32B}
& Negative & ` However', ` Therefore', ` But', `However', `Therefore', ` however', ` Thus', ` Hence', ` So', ` therefore' \\
\addlinespace
& Positive & `</', ` Gund', `(ed', ` perpetr', `/change', ` resil', `ted', `).\textasciicircum{}', ` grosse', ` accommod' \\
\midrule
\multirow{2}{*}{Qwen 3 32B}
& Negative & `\textcyrillic{профессионал}', \begin{CJK*}{UTF8}{gbsn}`但是'\end{CJK*}, `\textcyrillic{ зарегист}', `\textcyrillic{практик}', `\textcyrillic{слуша}', `\textcyrillic{чрежден}', `\textcyrillic{ формирова}', `\textcyrillic{ приятн}', `\textcyrillic{ кнопк}', \begin{CJK*}{UTF8}{gbsn}`但是如果'\end{CJK*} \\
\addlinespace
& Positive & ` gi\'{u}', \begin{CJK*}{UTF8}{gbsn}`勠', `到最后', `幌', `癞', `嗲', `佬', `匙', `总而言', `嗑'\end{CJK*} \\
\midrule
\multirow{2}{*}{Distill 32B}
& Negative & \begin{CJK*}{UTF8}{gbsn}`但', `但是', `然而'\end{CJK*}, ` But', \begin{CJK*}{UTF8}{gbsn}`但实际上', `而'\end{CJK*}, ` but', ` However', `/XMLSchema', \begin{CJK*}{UTF8}{gbsn}`所以'\end{CJK*} \\
\addlinespace
& Positive & \begin{CJK*}{UTF8}{gbsn}`大大小', `有的玩家'\end{CJK*}, `.SuspendLayout', \begin{CJK*}{UTF8}{gbsn}`挺好', `揠', `文章来源', `蠲', `帮我', `瘦身', `都喜欢'\end{CJK*} \\
\bottomrule
\end{tabular}
\end{table*}

\section{Additional Correlations}
We track how well verbalized confidence, $v_\text{conf}$, or $v_\text{commit}$ track binary answer commitment and $p_\text{bt}$ on the test set. For the linear directions, we compute the projections of the answer proposal hidden activations. We record the Spearman $\rho$ with answer commitment in \autoref{tab:marker-corr-commit} and $p_\text{bt}$ in \autoref{tab:marker-corr-pback}. 

\begin{table}[htbp]
\centering
\small
\setlength{\tabcolsep}{4pt}
\caption{\revise{Spearman $\rho$ against answer commitment, on the test set. $\rho_{\max}$: the largest $\rho$ attainable against this binary target at its class balance. $^{*}$: 95\% rollout-cluster bootstrap CI excludes 0.}}
\begin{tabular}{lll rrr}
\toprule
 & & & \multicolumn{3}{c}{Spearman $\rho$} \\
\cmidrule{4-6}
Model & Score & Fit on (site) & All & Correct & Incorrect \\
\midrule
\multirow{4}{*}{Olmo 3 32B} & Commit dir. & GPQA-D (43.attn) & {\boldmath$+0.79^{*}$} & {\boldmath$+0.79^{*}$} & {\boldmath$+0.71^{*}$} \\
 & Confidence dir. & AIME 2026 (23.attn) & $+0.49^{*}$ & $+0.51^{*}$ & $+0.36^{*}$ \\
 & Verbalized conf. & -- & $+0.33^{*}$ & $+0.40^{*}$ & $+0.09^{\phantom{*}}$ \\
 & \textit{$\rho_{\max}$} & & $0.87$ & $0.87$ & $0.86$ \\
\midrule
\multirow{4}{*}{Qwen 3 32B} & Commit dir. & GPQA-D (51.mlp) & {\boldmath$+0.82^{*}$} & {\boldmath$+0.81^{*}$} & {\boldmath$+0.78^{*}$} \\
 & Confidence dir. & AIME 2026 (47.attn) & $+0.64^{*}$ & $+0.70^{*}$ & $+0.58^{*}$ \\
 & Verbalized conf. & -- & $+0.26^{*}$ & $+0.41^{*}$ & $-0.03^{\phantom{*}}$ \\
 & \textit{$\rho_{\max}$} & & $0.87$ & $0.86$ & $0.86$ \\
\midrule
\multirow{5}{*}{Distill 32B} & Commit dir. & MMLU-Pro (46.attn) & {\boldmath$+0.79^{*}$} & {\boldmath$+0.76^{*}$} & {\boldmath$+0.79^{*}$} \\
 & Confidence dir. & GSM8K (46.attn) & $-0.00^{\phantom{*}}$ & $-0.01^{\phantom{*}}$ & $-0.03^{\phantom{*}}$ \\
 & Verbalized conf. & -- & $+0.22^{*}$ & $+0.23^{*}$ & $+0.00^{\phantom{*}}$ \\
 & \textit{$\rho_{\max}$} & & $0.87$ & $0.86$ & $0.84$ \\
\bottomrule
\end{tabular}
\label{tab:marker-corr-commit}
\end{table}

\begin{table}[htbp]
\centering
\small
\setlength{\tabcolsep}{4pt}
\caption{\revise{Spearman $\rho$ with the next-token probability of a backtrack token, on the test markers. $^{*}$: 95\% rollout-cluster bootstrap CI excludes 0.}}
\label{tab:marker-corr-pback}
\begin{tabular}{lll rrr}
\toprule
 & & & \multicolumn{3}{c}{Spearman $\rho$} \\
\cmidrule{4-6}
Model & Score & Fit on (site) & All & Correct & Incorrect \\
\midrule
\multirow{3}{*}{Olmo 3 32B} & Commit dir. & GPQA-D (43.attn) & $-0.69^{*}$ & $-0.71^{*}$ & $-0.62^{*}$ \\
 & Confidence dir. & AIME 2026 (23.attn) & $-0.68^{*}$ & $-0.69^{*}$ & $-0.61^{*}$ \\
 & Verbalized conf. & -- & $-0.32^{*}$ & $-0.38^{*}$ & $-0.05^{\phantom{*}}$ \\
\midrule
\multirow{3}{*}{Qwen 3 32B} & Commit dir. & GPQA-D (51.mlp) & $-0.63^{*}$ & $-0.62^{*}$ & $-0.63^{*}$ \\
 & Confidence dir. & AIME 2026 (47.attn) & $-0.60^{*}$ & $-0.61^{*}$ & $-0.58^{*}$ \\
 & Verbalized conf. & -- & $-0.22^{*}$ & $-0.29^{*}$ & $-0.15^{\phantom{*}}$ \\
\midrule
\multirow{3}{*}{Distill 32B} & Commit dir. & MMLU-Pro (46.attn) & $-0.65^{*}$ & $-0.65^{*}$ & $-0.53^{*}$ \\
 & Confidence dir. & GSM8K (46.attn) & $-0.46^{*}$ & $-0.45^{*}$ & $-0.46^{*}$ \\
 & Verbalized conf. & -- & $-0.28^{*}$ & $-0.32^{*}$ & $-0.09^{\phantom{*}}$ \\
\bottomrule
\end{tabular}
\end{table}

\end{document}